\documentclass{article}
\usepackage[preprint]{neurips_2025_custom}
\usepackage{fix-cm}

\definecolor{linkColor}{rgb}{0.2,0.4,0.6}
\usepackage[utf8]{inputenc}
\usepackage[T1]{fontenc}
\usepackage[colorlinks=true,linkcolor=linkColor,citecolor=linkColor,filecolor=linkColor,urlcolor=linkColor]{hyperref}
\usepackage{url}
\usepackage{graphicx}
\usepackage{microtype}

\usepackage{amsmath}
\usepackage{amssymb}
\usepackage{amsfonts}
\usepackage{array}
\usepackage{booktabs}
\usepackage{caption}
\usepackage[inline]{enumitem}
\usepackage{tabularx}
\usepackage{listings}
\usepackage{xcolor}

\newcolumntype{L}[1]{>{\raggedright\arraybackslash}p{#1}}
\newcolumntype{R}[1]{>{\raggedleft\arraybackslash}p{#1}}
\newcolumntype{Y}{>{\raggedright\arraybackslash}X}

\setlist[itemize]{leftmargin=*, itemsep=2pt, topsep=4pt}
\setlist[enumerate]{leftmargin=*, itemsep=2pt, topsep=4pt}
\definecolor{codebg}{rgb}{0.97,0.97,0.97}
\definecolor{codekw}{rgb}{0.18,0.42,0.60}
\definecolor{codecomment}{rgb}{0.45,0.45,0.45}

\lstdefinestyle{pgr}{
  basicstyle=\ttfamily\scriptsize,
  backgroundcolor=\color{codebg},
  keywordstyle=\color{codekw}\bfseries,
  commentstyle=\color{codecomment}\itshape,
  stringstyle=\color{codecomment},
  breaklines=true,
  showstringspaces=false,
  frame=none,
  numbers=none,
  xleftmargin=6pt,
  aboveskip=6pt,
  belowskip=4pt,
  columns=fullflexible,
}
\newcommand{\Realm}{\mathcal{R}}
\newcommand{\Space}{\mathcal{S}}

\title{post-graph-rag: A PostgreSQL-Native Bi-Temporal Graph RAG Engine with Temporal Grounding at Synthesis}
\author{%
Chandan Rajah\thanks{\texttt{chandan.rajah@gmail.com}} \\
Independent Researcher
}
\date{}

\begin{document}

\maketitle

\fancyhead[C]{\raisebox{-6pt}{\hspace{-32pt}\normalsize\bfseries
  post-graph-rag: A PostgreSQL-Native Graph RAG Engine}}

\begin{abstract}
\setlength{\parskip}{0.45\baselineskip}%
Graph RAG connects facts no single passage states, but implementations pay three times: in infrastructure, keeping vector store, graph database and document store in sync; in quality, because a pipeline that never refuses extractor output stores edges asserting nothing; and over time, because a graph that only accumulates treats superseded and current facts alike.

\texttt{post-graph-rag} is an open-source engine addressing all three. Chunks with embeddings, a canonical entity graph and community summaries live in one PostgreSQL database, with \texttt{pgvector} for search and edge tables for traversal. Extraction output is validated before writing: vague predicates, pronominal names and bare quantities are rejected, predicates normalise onto an optional vocabulary, and entities resolve to one vertex per canonical name. A bi-temporal layer records when a relation held and when the system believed it, superseding incompatible earlier assertions from document order.

Against LightRAG on three corpora with extraction and embedding models fixed, it builds a denser and more queryable graph, and supersedes relationships where a baseline with no temporal model supersedes none.

On LongMemEval \citep{wu2024longmemeval}, 500 questions of long-horizon chat memory, it scores 94.0\% with \texttt{gemini-3.6-flash} against 71.2 for Zep's \texttt{gpt-4o} \citep{rasmussen2025zep} and 60.2 for full context, leading on all six question types. The largest single contribution is temporal grounding in the prompt: carrying each relation's validity period through to synthesis moves temporal reasoning from 0.496 to 0.881, ablated paired on one graph per instance. On ECT-QA, earnings-call transcripts restating every metric each quarter, it scores 0.807 under that benchmark's own protocol against 0.599, 0.406 and 0.405 published for TG-RAG, LightRAG and GraphRAG.

\medskip
\noindent\textbf{Code.} \href{https://github.com/crajah/post-graph-rag}{\texttt{post-graph-rag}} is the engine; \href{https://github.com/crajah/post-graph}{\texttt{post-graph}} is the PostgreSQL graph layer it builds on. Both are Apache 2.0 licensed and on PyPI.
\end{abstract}

\section{Introduction}

Retrieval-augmented generation grounds a language model in retrieved text \citep{lewis2020rag,karpukhin2020dpr}, and remains the default way to put a model to work over a private corpus. Its dominant failure mode is well documented and structural rather than incidental: retrieval returns passages ranked by similarity to the question, so any answer that requires combining facts stated in two different passages --- particularly two passages in different documents that share no vocabulary --- is not retrievable at all \citep{gao2023ragsurvey,barnett2024failure}. No amount of re-ranking fixes this, because the evidence needed to rank the second passage highly is only visible after the first has been read.

Graph RAG addresses this by extracting a knowledge graph from the corpus at index time and retrieving over both the graph and the text \citep{edge2024graphrag,guo2024lightrag,gutierrez2024hipporag,peng2024graphragsurvey}. Entities mentioned in several documents become one vertex, so traversal reaches facts that lexical similarity never would; clustering the graph and summarising each cluster additionally makes corpus-level questions (``what are the main themes here?'') answerable, which passage retrieval cannot do because no passage contains the answer \citep{edge2024graphrag}.

Three costs have kept this from being routine.

\paragraph{Infrastructure.} A typical Graph RAG deployment runs a vector database, a graph database and a document store side by side, and the application becomes responsible for keeping three systems consistent under partial failure, for authorising three sets of credentials, and for backing up three data stores whose snapshots are not mutually consistent. For teams already running PostgreSQL --- which is most teams --- this is a large amount of new infrastructure to justify.

\paragraph{Graph quality.} LLM-based triple extraction \citep{trajanoska2023kgllm,zhu2023llmkg} is a generation task, and a generative pipeline with no rejection criterion always produces output. In practice much of that output is unusable as graph structure: predicates such as \texttt{relates\_to} record that two entities co-occurred and nothing more; pronouns and relative references (\texttt{he}, \texttt{his father}) become vertices that cannot resolve or connect; the same entity fragments across \texttt{Babbage}, \texttt{Charles Babbage} and \texttt{C.\ Babbage}; bare figures such as \texttt{\$18.4 billion} arrive as entities; and the same triple re-extracted from an overlapping chunk becomes a second parallel edge. Each is individually minor and collectively fatal, because a graph in which half the edges assert nothing is a graph whose traversals return noise --- noise that is expensive, since every retrieved triple consumes synthesis context.

\paragraph{Time.} A knowledge graph built by accumulation is implicitly a claim that everything in it holds simultaneously. On any corpus whose facts change --- a novel sequence in which alliances reverse, a decade of annual filings in which a programme goes from generating record cash flow to consuming it --- that claim is false, and the failure is not a retrieval miss but a wrong answer stated confidently. Asked what two characters are to each other, a purely accumulative graph reports that they are allies \emph{and} enemies. The information needed to resolve this is usually in the corpus, but it is in the \emph{ordering} of the documents rather than in any one of them.

\bigskip
\noindent This paper describes \texttt{post-graph-rag}, an open-source engine that attacks all three.\footnote{\url{https://github.com/crajah/post-graph-rag}, Apache 2.0 licensed.} The entire dual representation lives in one PostgreSQL database, with \texttt{pgvector} \citep{pgvector} providing HNSW approximate nearest-neighbour search \citep{malkov2020hnsw} and vertex/edge tables providing traversal. Against that substrate we place extraction-time invariants that decide what is allowed to become graph structure at all, and we make the pipeline \emph{fail closed}: when extraction produces nothing usable, indexing raises rather than writing a placeholder, because once written a placeholder edge is indistinguishable from a genuine one. On top of that we add a temporal layer that records when a relation held and lets a later document close an earlier, incompatible one.

Our contributions are:

\begin{enumerate}
\item A \textbf{single-store Graph RAG architecture} in which vector passages, a canonical entity graph, document-to-entity mention edges and community summaries are co-located in PostgreSQL, with multi-tenancy expressed as a two-level \emph{realm}/\emph{space} partition and schema-level isolation per tenant (Section~\ref{sec:design}).
\item A set of \textbf{extraction-time invariants} --- predicate normalisation and controlled-vocabulary snapping, rejection of vague, pronominal, conjunctive, over-long and purely quantitative names, alias-based entity resolution with a longest-form canonicalisation rule, provenance-counted corroboration, and explicit negation --- together with the fail-closed policy that makes them meaningful (Section~\ref{sec:extraction}).
\item A \textbf{bi-temporal relation model}: optional validity intervals extracted only when the text states them, belief time recording when the system learned and unlearned each relation, supersession of incompatible relations resolved from document order rather than from model-supplied dates, \texttt{as-of} retrieval against either axis, and idempotent re-indexing in which entities losing their last mention go dormant rather than being deleted (Section~\ref{sec:temporal}).
\item A \textbf{retrieval design} in which relations reach the prompt by three fused channels --- bounded traversal, relation-embedding similarity and lexical search --- with denied relations rendered as denials and each relation carrying its validity period into the prompt. That last step, temporal grounding at synthesis, is the largest single contributor we measure: ablating it costs 38.4 points on temporal reasoning (Sections~\ref{sec:retrieval} and~\ref{sec:validity}).
\item \textbf{Capabilities for exploration-first consumers}: a nested topic tree built by recursive supergraph clustering, retrieval-coverage telemetry that identifies never-retrieved regions of a corpus, a belief-time delta API with database-clock watermarks, and retention expressed as reversible demotion rather than deletion (Section~\ref{sec:exploration}).
\item An \textbf{evaluation} in two registers: end-to-end on LongMemEval, where the engine scores 94.0\% and leads the strongest published Zep/Graphiti configuration on all six question types, and on ECT-QA; and mechanism-level --- per-mechanism ablations, extraction-model sensitivity across four models, and a LightRAG comparison on three corpora under identical extraction and embedding models. The paired-ablation protocol these use, holding one graph fixed across arms, is itself a finding: re-indexing alone moves this benchmark further than most effects worth testing (Section~\ref{sec:evaluation}).
\end{enumerate}

We are explicit about scope. The corpus comparisons are engineering measurements with single runs per configuration; the benchmark result is a single run under one protocol, and what it supports is bounded accordingly. Section~\ref{sec:limitations} bounds what each supports.

\section{Related Work}

\paragraph{Retrieval-augmented generation.} \citet{lewis2020rag} introduced RAG as joint training over a retriever and a generator; dense passage retrieval \citep{karpukhin2020dpr} established the bi-encoder pattern that most production systems still use. Surveys \citep{gao2023ragsurvey} and failure analyses \citep{barnett2024failure} identify multi-hop and corpus-level questions as the class RAG systematically misses, and evaluation frameworks such as RAGAS \citep{es2023ragas} score retrieval faithfulness without addressing whether the required evidence was co-locatable in the first place.

\paragraph{Graph-based RAG.} \citet{edge2024graphrag} established the pattern this work follows most closely: extract entities and relations with an LLM, cluster the entity graph, summarise each cluster, and answer global questions from the summaries. LightRAG \citep{guo2024lightrag} is our direct baseline and the source of the dual-level keyword decomposition --- low-level keywords naming concrete entities, high-level keywords naming themes --- which we adopt for retrieval conditioning. It stores its graph in an in-process NetworkX object with a file-backed vector index by default, and labels edges with free-text keywords; Section~\ref{sec:evaluation} shows what each of those choices costs. HippoRAG \citep{gutierrez2024hipporag} runs personalised PageRank over the extracted graph rather than clustering it, and RAPTOR \citep{sarthi2024raptor} obtains hierarchical abstraction with no graph at all. \citet{peng2024graphragsurvey} survey the area. That structure alone does not pay is now measurable: \citet{rusu2026forgetting} build a per-turn conversational knowledge graph that \emph{loses} to a flat vector baseline on LongMemEval overall (LLM-judge 0.454 against 0.536), winning only temporal reasoning --- and their diagnosis of why reads as a design checklist this system already follows: graph extraction that replaces the text discards what verbatim recall needs (we retain chunks and reach them through mention edges), and conflict resolution that keeps an earlier attribute wants a last-write-wins policy (our supersession from document order). Our departure from all of these is the write path: we treat extraction output as untrusted input to be validated, we co-locate the resulting structure with the vectors, and we model when a relation held.

\paragraph{LLMs as knowledge-graph constructors.} \citet{trajanoska2023kgllm} and \citet{zhu2023llmkg} evaluate LLMs for knowledge-graph construction and report the failure modes that motivate our validation gates: unstable predicate vocabularies, unresolved coreference, and hallucinated relations between co-occurring entities. \citet{ji2022kgsurvey} survey representation and acquisition more broadly. The entity resolution we perform with model-supplied aliases plus a canonicalisation rule is a restricted case of collective entity resolution \citep{bhattacharya2007collective}; the restriction --- resolution within a tenant partition, by name and alias only --- is what makes it cheap enough to run inline during indexing. Explicit negation representation has a long history in clinical information extraction, where NegEx \citep{chapman2001negex} established that asserting and denying the same relation must be distinguishable downstream; we make the distinction structural rather than lexical.

\paragraph{Time in databases and knowledge graphs.} Valid-time modelling --- recording the interval during which a fact holds in the world, separately from when it was recorded --- is long-established in temporal databases \citep{jensen1999temporal}. Temporal knowledge-graph work has largely approached the problem as representation learning, embedding timestamped quadruples for link prediction and completion \citep{dasgupta2018hyte,garciaduran2018sequence}. Our setting is different in a way that matters: we are not completing a graph whose timestamps are given, we are \emph{constructing} one from prose that mostly does not state dates at all. Two consequences follow. Validity intervals are populated only when the text states them, and relations without one are treated as always valid rather than as never valid --- silence about when a fact held means it held throughout. And conflict resolution is driven by \emph{document order}, not by extracted dates, because document order is available on every corpus whereas a stated period is available on few. Closest to our setting is Zep \citep{rasmussen2025zep}, whose Graphiti engine is also bi-temporal, separating when a fact held from when the system learned it, and which we adopt three retrieval mechanisms from (Section~\ref{sec:diversification}). Two differences remain. Zep invalidates an edge when an LLM judges a new one to contradict it, whereas our primary mechanism is declarative --- a deployment names sets of predicates that cannot co-hold, and supersession then follows from document order without a model call --- with LLM contradiction detection available as a complement rather than as the mechanism. And Zep runs on a dedicated graph service, where we target the PostgreSQL instance an application already operates. To our knowledge the specific combination of extraction-time supersession from declared exclusive predicate groups, resolved by document order and layered over optional valid-time intervals, is not offered elsewhere.

\paragraph{Community detection.} Cluster quality determines whether a ``theme'' summary describes a theme or describes everything. We use Leiden \citep{traag2019leiden}, which repairs the badly-connected-community defect of Louvain \citep{blondel2008louvain}, and fall back to deterministic label propagation \citep{raghavan2007label} where the native build is unavailable. Determinism is load-bearing: a randomised partition would produce a different set of stored summaries on every indexing run.

\paragraph{PostgreSQL as a substrate.} \texttt{pgvector} \citep{pgvector} adds vector columns and HNSW indexes \citep{malkov2020hnsw} to PostgreSQL. Graph workloads in PostgreSQL have been approached through Apache AGE \citep{apacheage}, which embeds an openCypher engine. We take the lighter path: relations are ordinary edge tables with foreign keys and JSONB payloads, and traversal is bounded $k$-hop SQL expansion from vector-matched seeds (two hops by default, with temporal predicates applied within the walk) --- sufficient because Graph RAG retrieval is dominated by shallow expansion around matched entities rather than by deep path queries. The substrate additionally exposes an openCypher surface that compiles to the same SQL, so declarative path queries are available to callers who want them without the retrieval path depending on a query engine, and hot attributes --- identifiers and validity bounds --- are promoted from JSONB to generated columns that indexes can address directly.

\section{System Design}
\label{sec:design}

\subsection{Overview}

\begin{figure}[t]
  \centering
  \includegraphics[width=0.96\linewidth]{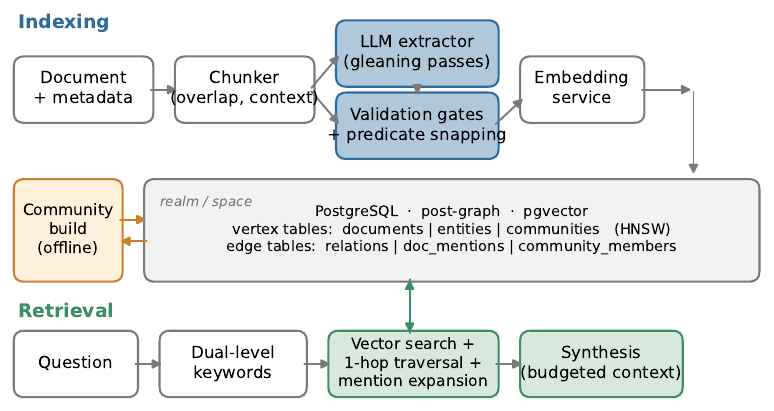}
  \caption{System architecture. What a conventional deployment splits across three systems --- vector index, graph, document store --- together with the derived community summaries, is one set of tables in a single PostgreSQL database, partitioned by realm and space. The LLM extractor and the validation gates are separate stages by design: the extractor is a generative component and is treated as untrusted, while the gates are deterministic and decide what becomes graph structure.}
  \label{fig:architecture}
\end{figure}

Figure~\ref{fig:architecture} shows the two paths through the system. Indexing chunks a document, extracts entities and triples with an LLM, validates that output, embeds chunks and entities, and writes vertices and edges. Retrieval embeds the question, then gathers relations by three fused channels --- bounded traversal from vector-matched entities, similarity search over relation embeddings, and lexical search --- pulls in chunks mentioning the matched entities, optionally retrieves community summaries, and synthesises an answer, with each relation carrying its validity period into the prompt. Community construction is a third, offline path that reads the entity graph and writes summaries back into the same database.

\subsection{Data model}

Vertex and edge tables, their audit and history shadows, and the realm/space partitioning below are provided by \texttt{post-graph}, a separate PostgreSQL-backed graph library.\footnote{\url{https://github.com/crajah/post-graph}, Apache 2.0 licensed. Splitting it out keeps the graph substrate usable --- and testable --- independently of anything RAG-specific.} Against it, the engine defines three vertex tables and three edge tables (Table~\ref{tab:schema}). Vertex tables carry a JSONB \texttt{payload} and a \texttt{vector(d)} embedding column indexed with HNSW; edge tables carry endpoint identifiers, a \texttt{relation\_type} string, and a JSONB payload. Documents are chunks rather than whole files: the chunk is the unit that is embedded, retrieved and cited, and it carries the structured provenance (\texttt{source}, \texttt{category}, \texttt{collection}, \texttt{document}, \texttt{page}, \texttt{paragraph}) needed to render a citation, plus a stable \texttt{doc\_key} and \texttt{content\_hash} used for idempotent re-indexing (Section~\ref{sec:reindex}).

\begin{table}[t]
\centering
\small
\caption{Tables provisioned per realm. Vertex tables carry an HNSW-indexed embedding column; edge tables carry endpoint identifiers and a JSONB payload. Relation embeddings are enabled by default as of version 1.5.0: Section~\ref{sec:threechannel} shows that traversal from a vector-matched entity cannot reach a relation whose endpoints are generically named, and that no re-ranking recovers it.}
\label{tab:schema}
\begin{tabular}{@{}llL{6.1cm}@{}}
\toprule
\textbf{Table} & \textbf{Kind} & \textbf{Contents} \\
\midrule
\texttt{documents} & vertex & Text chunk, \texttt{DocumentMetadata}, document key, content hash, embedding \\
\texttt{entities} & vertex & Canonical name, type, description, alias list, dormancy stamp, embedding \\
\texttt{communities} & vertex & Cluster report (title, summary, findings, rating), build time, embedding \\
\texttt{relations} & edge & entity $\to$ entity; predicate, description, contributing chunks, weight, negation flag, confidence, validity interval, supersession pointer \\
\texttt{doc\_mentions} & edge & chunk $\to$ entity mentioned in it \\
\texttt{community\_members} & edge & community $\to$ member entity \\
\texttt{community\_children} & edge & community $\to$ child community, when a hierarchy is built \\
\bottomrule
\end{tabular}
\end{table}

\subsection{Two-level tenancy}

Rows are partitioned along two axes. A \emph{realm} $\Realm$ is a tenant: an independent knowledge base with its own entity namespace. A \emph{space} $\Space$ is an application-level sub-grouping within a realm --- \texttt{production}, \texttt{sandbox}, a per-user workspace --- and both indexing and retrieval are scoped to it. The distinction matters because entity identity is defined per space: entities are unique per $(\Realm, \Space, \operatorname{lower}(\text{name}))$, enforced by a unique index that both guarantees the invariant and makes canonical-name lookup an index hit rather than a scan. A query may set \texttt{space} to a reserved wildcard to read across all spaces, but the wildcard is a read-time filter only and is rejected on writes, since an entity written to ``all spaces'' would have no defined identity.

Space scoping is applied to traversal as well as to vector search. Scoping only the search is a subtle leak: the seed entity is correctly confined to the caller's space, but a one-hop expansion from that seed then walks into edges and vertices belonging to another space, and the leaked facts arrive in the synthesis context indistinguishable from the caller's own.

Realms may share one physical set of tables filtered by a \texttt{realm} column, or --- recommended, and used for all measurements here --- receive a PostgreSQL schema each. With shared tables the first realm to create the \texttt{entities} table fixes the embedding column width for every subsequent realm, so a second realm using a different embedding model simply cannot be created. Schema-per-realm also makes tenant deletion a \texttt{DROP SCHEMA} rather than a cascade of deletes.

\subsection{Concurrency and write ordering}
\label{sec:concurrency}

Indexing is almost entirely network-bound: per chunk, one or more extraction calls and one batched embedding call. Chunks are therefore processed in batches whose LLM and embedding work runs concurrently, while the resulting \emph{graph writes are applied in order}.

The asymmetry is deliberate. Entity resolution is a read-modify-write against a uniqueness index --- look up by canonical name, then by alias, merge, upsert --- and concurrent writers racing on the same name split entities that should have merged. Serialising the write phase is what makes \texttt{Babbage} and \texttt{Charles Babbage} reliably converge on one vertex under concurrency. Batching also preserves coreference context at batch granularity: each batch sees the canonical entity names discovered by earlier batches, so a chunk from the middle of a document still has something to resolve its references against. Setting the batch size to 1 recovers strict sequential indexing with per-chunk context.

This costs indexing throughput relative to a system that parallelises across documents into an in-process graph object with no uniqueness constraint to respect. It is the price of an entity graph that is correct under concurrency, paid once at index time; Section~\ref{sec:lightrag} explains why we do not quantify it as a comparison.

\subsection{Fail-closed invariants}
\label{sec:failclosed}

Three classes of failure are made loud rather than silent, on the principle that a failure invisible at write time and expensive at read time is worse than an exception.

\emph{Missing vector support.} If the \texttt{vector} extension is unavailable when a vertex table is created, the table is created without an embedding column and every subsequent similarity search returns nothing --- writes succeed, and retrieval quietly degrades to empty. Schema initialisation therefore verifies, for every vertex table, that an \texttt{embedding} column exists and that its declared width equals the configured embedding dimension, and raises otherwise. The same check catches a table created earlier under a different embedding model.

\emph{Unusable extraction.} If the LLM returns no usable entities or triples for a chunk, or if every candidate is rejected by the gates of Section~\ref{sec:gates}, that chunk raises. An individual bad chunk is skipped and reported, but a run in which \emph{every} chunk fails raises rather than completing, so a total provider outage cannot be mistaken for having successfully indexed an empty corpus. Synthesising a placeholder edge so the pipeline continues is never an option: a placeholder is indistinguishable from genuine extracted structure once stored, so a transient outage during a long run would permanently poison the graph.

\emph{Embedding failure.} A locally computed fallback vector is not comparable with an API embedding; the two live in unrelated geometries, so distances between them are meaningless. Mixing them into one index corrupts retrieval rather than degrading it. Embedding failures therefore raise by default, and the fallback path is opt-in; when enabled, the deterministic fallback uses SHA-256 for both bucket assignment and magnitude rather than Python's built-in \texttt{hash}, which is salted per process --- with the salted variant, nothing indexed in one process could ever be retrieved in another.

Retries are bounded by a wall-clock deadline per call rather than by attempt count alone. Without one, a sustained outage costs retries $\times$ models $\times$ backoff on \emph{every} call; a twelve-community build once spent 38 minutes producing nothing.

\section{Extraction}
\label{sec:extraction}

\subsection{Chunking with document context}

Documents are split into overlapping chunks. Overlap matters specifically for graph extraction rather than for retrieval: a relation whose subject appears in the last sentence of one chunk and whose object appears in the first sentence of the next is invisible to both without it. The default splitter packs whole paragraphs up to a character budget, carries a fixed overlap forward, and resumes the overlap at a sentence boundary so it reads as prose to the extractor. Callers may substitute any splitter.

More consequentially, a \emph{document context} is threaded through the chunks of a document: its title, its source, and the canonical names of the entities identified so far, capped at a configurable limit. Without it every chunk after the first is extracted blind, and its pronouns and definite references have nothing to resolve against --- which is precisely how a vertex named \texttt{his father} enters a graph. With it, the extractor is instructed to reuse the exact canonical names it is given.

\subsection{Gleaning}

Single-pass extraction reliably under-recalls on dense prose. After the first pass, the extractor re-prompts with the entities and triples already found and asks only for what was missed, merging the result. Each pass costs one additional LLM call per chunk; Section~\ref{sec:ablation} measures the recall it buys. Gleaning is an enhancement rather than a requirement --- a failed gleaning pass is logged and the earlier result kept, because the first pass has already succeeded and discarding it would trade a partial result for none.

\subsection{Validation gates}
\label{sec:gates}

Every extracted record passes through deterministic gates before it can become graph structure (Table~\ref{tab:gates}). The gates are stated as rejections rather than repairs: a record that cannot be made into a well-formed assertion is dropped, not guessed at.

\begin{table}[t]
\centering
\small
\caption{Validation gates applied to extractor output. Each rejects a class of record that is well-formed as text but meaningless as graph structure. The possessive gate is off by default: on real prose it also rejects legitimate named things (``Amp\`ere's force law''), so roughly half its hits are false positives.}
\label{tab:gates}
\begin{tabular}{@{}L{2.85cm}L{4.4cm}L{4.5cm}@{}}
\toprule
\textbf{Gate} & \textbf{Rejects} & \textbf{Why it is not repairable} \\
\midrule
Vague predicate & \texttt{relates\_to}, \texttt{associated\_with}, \texttt{see\_also}, \dots & An edge so labelled is indistinguishable from no edge; the specific relation is not recoverable from the label \\
Pronominal name & \texttt{he}, \texttt{they}, \texttt{his father}, \texttt{the company} & The referent varies by chunk, so the vertex can never resolve or connect \\
Conjunctive name & ``Ada Lovelace and Charles Babbage'' as one entity & Two entities, not one; matches only a bare conjunction of proper-noun phrases, so titles containing ``and'' survive \\
Bare quantity & \texttt{\$18.4 billion}, \texttt{2.5\%}, \texttt{\$1,326} & A figure is the value of a relation, not a thing that holds relations; every filing reports different figures, so such vertices never recur \\
Over-long name & names beyond 200 characters & A whole table row returned as a name overflows the unique btree index and aborts the run \\
Possessive role phrase & ``Babbage's father'' \emph{(opt-in)} & Names a role, not a nameable entity \\
Self-loop & subject $=$ object & Asserts nothing about a pair \\
Confidence floor & relations below a configured threshold & Hedged assertions stored as facts \\
Orphaned endpoint & triples whose endpoint was itself rejected & Would reintroduce the rejected name as a stub vertex \\
\bottomrule
\end{tabular}
\end{table}

The predicate gate is the load-bearing one. Prohibiting vague connectors is what enforces the rule that two entities appearing near each other never produces an edge: if the text does not state a specific relationship, the extractor is instructed to emit no triple for the pair, and any co-occurrence edge it emits anyway is rejected here.

\subsection{Entity granularity}
\label{sec:granularity}

A gate can only reject a name that is malformed. A subtler failure --- one we found only on financial prose --- is a name that is well-formed but not \emph{stable}. Asked to extract from an annual report, models readily mint filing-specific compound entities:

\begin{quote}\ttfamily\small
Boeing Commercial Airplanes revenue increase\\
Boeing Company Q4 2005 net loss\\
737 programme production impacts
\end{quote}

Each is a legitimate noun phrase and passes every gate. But none can ever recur in another document, so retrieval for ``cash flow'' spreads across dozens of hyper-specific vertices instead of landing on a few well-populated ones, and supersession --- which only fires when the same entity pair is characterised twice --- has nothing to work with.

The fix is a prompt constraint rather than a gate, because the judgement is semantic: emit the stable entity that could plausibly be named again in a \emph{different} document on the same subject, and put the movement, period and magnitude in the relation and its description instead. The test the extractor is asked to apply before emitting a name is whether that exact name could appear in another document about this subject. Section~\ref{sec:finance} measures the effect: retrieval recall rose between $2\times$ and $9\times$ on the affected questions.

\subsection{Predicate normalisation and controlled vocabulary}

Surviving predicates are normalised: lowercased, underscore-joined, and stripped of leading tense auxiliaries, so \texttt{was\_appointed\_knight\_of} and \texttt{appointed\_knight\_of} collapse onto one predicate. Left at that, extraction produces roughly one predicate per relation --- maximally faithful to the text and useless for querying by relation type.

An optional controlled vocabulary addresses this in two stages. The vocabulary is rendered into the extraction prompt, steering the model towards it; extracted predicates are then \emph{snapped} onto it by prefix match, so \texttt{design} and \texttt{designs} both reach \texttt{designed}. A separate explicit alias map merges genuinely different wordings that no morphological rule relates (\texttt{collaborated\_with} $\to$ \texttt{worked\_with}). Inverse relations are deliberately \emph{not} merged onto their converse: mapping \texttt{educated\_by} onto \texttt{taught} would silently reverse the direction of every edge it touched. A long tail of single-use predicates survives by design --- \texttt{funded}, \texttt{lived\_at}, \texttt{co\_authored} are genuinely distinct relations a compact vocabulary does not cover --- and the commonest off-vocabulary predicates are reported by the analysis tooling, which is the feedback loop for extending it.

\subsection{Entity resolution}

Entities are resolved to one vertex per $(\Realm, \Space, \operatorname{lower}(\text{name}))$, and additionally through aliases. The extractor is required to record every other surface form it sees for an entity, so \texttt{Babbage}, \texttt{Charles Babbage} and \texttt{C.\ Babbage} converge. Resolution proceeds by canonical name first, then by alias; on a hit, the two records are merged under a policy in which a more specific value never loses to a placeholder:

\begin{itemize}
\item The \textbf{longer surface form becomes canonical} and the shorter is demoted to an alias, so \texttt{Babbage} upgrades to \texttt{Charles Babbage} and never the reverse.
\item A bare \texttt{Concept} type never overwrites a specific type, and a longer description never loses to a shorter one. The engine creates bare \texttt{Concept} stubs for triple endpoints the extractor did not return as full entities; without this rule, a stub arriving after a rich record would strip it.
\item Alias sets are unioned, minus any alias equal to the new canonical name.
\end{itemize}

A concurrent writer inserting the same name loses the unique-index race and falls back to resolving against the row that won, rather than failing the chunk. This is what makes cross-document traversal work at all: the same entity mentioned in many documents is one vertex, which is exactly what lets the graph connect chunks that share no vocabulary.

\subsection{Corroboration and negation}

The same triple extracted from several chunks --- routine, given overlap --- becomes one edge rather than several parallel edges. Its \texttt{weight} counts the \emph{distinct contributing chunks}, recorded as provenance on the edge, not the number of times it was written. The distinction matters for idempotence: incrementing blindly meant that re-indexing a document made its claims look independently corroborated. Weight is then a genuine corroboration signal, used to break ties when ranking relations for retrieval.

Relations the text explicitly denies are stored with the \emph{positive} predicate and a \texttt{negated} flag, not as an inverted predicate such as \texttt{did\_not\_have\_relationship\_with}. An inverted predicate is a new, unrelated vocabulary entry: a traversal filtering on \texttt{worked\_with} would not see it, and a summariser reading it would have to parse English negation out of an identifier. With the flag, clustering can down-weight denied edges, synthesis can render them explicitly as denials, and a consumer that does not care can drop them with a single predicate.

\section{The Temporal Model}
\label{sec:temporal}

\subsection{Validity intervals}

A relation may carry \texttt{valid\_from} and \texttt{valid\_to}, populated \emph{only} when the text states a period. Extracted dates are normalised to a bare year, year-month or full date; anything vaguer (``later'', ``in his youth'') is discarded rather than coerced, on the grounds that a wrong date is worse than no date. Partial dates are padded for comparison so \texttt{1625} orders below \texttt{1625-06-01}.

A relation with no stated period is treated as valid at every point in time. This is the load-bearing default: silence about when a fact held means it held throughout, not that it held nowhere, so a corpus that never mentions dates behaves exactly as it would without the temporal layer. Re-observing a relation without a period never erases a period recorded earlier.

\subsection{Supersession from document order}

Validity intervals only help where the prose supplies dates, which on most corpora is rarely. The mechanism that does the work instead is \emph{supersession}.

A deployment declares \emph{exclusive predicate groups} --- sets of predicates that cannot simultaneously hold between the same ordered pair. On a novel sequence, $\{$\texttt{friend\_of}, \texttt{enemy\_of}, \texttt{rival\_of}, \texttt{ally\_of}, \texttt{opponent\_of}$\}$ is one such group; on annual filings, $\{$\texttt{generates\_cash\_flow}, \texttt{consumes\_cash}$\}$ and $\{$\texttt{grounds}, \texttt{certifies}$\}$ are two of eight. When a new relation lands whose predicate shares a group with an existing relation on the same pair, the earlier edge is marked superseded by the newer one.

Two properties make this practical. Resolution is by \textbf{document order}, not by dates --- documents are indexed in a caller-supplied order (publication order for a novel sequence, fiscal year for filings), and the later assertion wins. This deliberately avoids depending on the model to state a period, which it does only when the text happens to say so. And superseded edges are \textbf{marked, never deleted}: the history is the point, and it remains reachable by asking for it explicitly.

\subsection{Belief time}
\label{sec:belieftime}

Validity records when a fact held in the world. It does not record when this system came to think so, and the two diverge whenever a corpus is indexed after the fact or revised later. Every relation therefore also carries a transaction-time pair, \texttt{t\_created} and \texttt{t\_expired}: the instant the relation entered the graph, and the instant belief in it was retracted. Supersession sets \texttt{t\_expired} on the edge it closes rather than deleting it, so the superseding edge and the superseded one are distinguishable by \emph{when the system stopped believing the latter}, independently of the period either describes.

This makes the model bi-temporal in the sense established for temporal databases \citep{jensen1999temporal} and adopted for agent memory by \citet{rasmussen2025zep}. The two axes answer different questions and neither substitutes for the other: valid time answers ``what was true in 2019'', belief time answers ``what would this system have told you in March, before the restatement was indexed''. The second is the auditable one --- it reconstructs a past answer rather than a past world --- which is what a deployment needs when it has to explain a decision it made on the evidence it had.

Stamps are ISO-8601 UTC so they order lexically without parsing. A relation with no \texttt{t\_created} predates the field and is treated as always known, on the same principle as an absent validity period: the absence records that nothing was stated, not that the answer is no.

\subsection{As-of retrieval}

A query may carry an \texttt{as\_of} date, in which case retrieved relations are filtered to those valid at that date, with undated relations always matching. A separate as-of-belief filter selects the graph's state of belief at an instant, admitting relations created at or before it and not yet retracted. Independently, superseded relations are excluded by default and can be requested. The filters compose: an \texttt{as\_of} query against a corpus with both mechanisms active returns the relations that a reader of the documents up to that point would have believed.

\subsection{Idempotent re-indexing and dormancy}
\label{sec:reindex}

A corpus that changes is also a corpus that gets re-indexed, and a graph built by appending turns a refresh into duplication. Each chunk therefore carries a stable \texttt{doc\_key} --- the source URL or path where available, the title otherwise, because a URL survives renaming --- and a content hash. Re-indexing a document whose chunk hashes are unchanged is a no-op. Where they differ, the document's previous chunks and their mention edges are removed and the document is re-extracted, so its contribution is \emph{replaced} rather than added to.

Removal raises a question about what happens to what the document taught. Entities are never deleted here: one whose last mentioning document disappears is marked \emph{dormant}, excluded from retrieval and from community rebuilds, and revived automatically if a later document mentions it again. Relations left with no contributing chunk are treated the same way. Deleting them would discard exactly the history the audit and append-only tables exist to keep.

Dormancy tracks document mentions only, and is deliberately independent of supersession: a superseded relation still evidences that its endpoints exist, so it must never push an entity dormant.

\section{Communities}
\label{sec:communities}

Corpus-level questions are answered from summaries of clustered subgraphs rather than from retrieved passages. After indexing, the engine snapshots the entity graph for a space, forms an undirected weighted graph in which each relation contributes its corroboration weight scaled by a factor $<1$ if it is a denial, and partitions it with Leiden \citep{traag2019leiden}, falling back to deterministic label propagation \citep{raghavan2007label} where the native build is unavailable. Clusters below a minimum size are dropped, since a two-entity ``community'' produces a report that says nothing its members' own descriptions do not.

Each surviving cluster is rendered as its member entities and the relations internal to it, and summarised by an LLM into a report with a title, an executive summary, a list of findings, and an importance rating. Denied relations are rendered as denials with an instruction never to report them as holding. The report is flattened to text, embedded, and stored as a vertex, with \texttt{community\_members} edges back to its entities --- so a report is always traceable to the subgraph that produced it, and global retrieval can find themes by vector similarity rather than by enumerating relations.

\paragraph{Ranking by more than similarity.} Retrieving community reports by pure vector similarity has a specific failure mode we observed directly: on ``what are the main themes?'' the engine drifted to a niche cluster about steampunk alternate histories, because a narrow cluster's tightly-worded summary can sit closer to a short, broad query than a wide cluster's summary does. Reports are therefore ranked on a blend of similarity, the report's own importance rating, and cluster size, with candidates over-fetched before re-ranking. Setting the importance and size weights to zero recovers pure similarity.

\paragraph{Rebuild, not accumulate.} Communities are derived data. A build clears the space's existing communities before writing new ones; otherwise a re-index leaves stale clusters describing a graph that no longer exists. Each report records when it was built, and retrieval warns when the oldest report predates the most recent graph write --- a derived artefact silently describing a superseded graph is exactly the class of error the rest of the system is built to make loud.

\paragraph{Title disambiguation and partial failure.} Reports are generated independently, so two clusters can be handed the same title; duplicates are qualified with a member entity not already named in the title. One unusable report is logged and skipped rather than abandoning the build, and clusters are processed largest first so a truncated build still covers the most significant structure. Global retrieval degrades to relation ranking when no communities have been built, so the mode never hard-fails.

\subsection{Structure, coverage and change}
\label{sec:exploration}

Three capabilities sit above the retrieval path and serve consumers that decide
\emph{what to look at next} rather than answer a question put to them. Each is
one engine call; none introduces an agent loop into the library.

\textbf{Hierarchy.} Communities are single-level by default. Setting a level
count builds a topic tree above them by recursive supergraph clustering: one
node per cluster, edge weights summed across the cut, re-clustered per level.
Recursion is what guarantees nesting --- a resolution ladder over the original
graph carries no such guarantee, and a hierarchy whose children are not
contained in their parents poisons every drill-down built on it. Parent reports
are synthesised from child \emph{reports} rather than from raw relations, which
caps the added LLM cost at roughly the cluster count per level.

\textbf{Coverage.} Optional telemetry records which entities and communities
each query touched, storing a hash of the query rather than its text. On it rest
two questions a corpus cannot otherwise answer about itself: which communities
have received least retrieval attention, and which entities have never been
retrieved at all. Telemetry is the one declared exception to the fail-closed
rule of Section~\ref{sec:failclosed}: read-side bookkeeping must never fail the
query it describes, so a failed event write is logged and swallowed.

\textbf{Change.} A delta reader answers what moved since an instant, in belief
time: relations added, relations superseded and by which, entities newly
dormant or revived, documents added, and whether communities are stale. The
watermark is the database clock rather than the caller's, which is what makes a
polling chain exactly-once under clock skew. Re-indexing an unchanged document
produces an empty delta, which turns the idempotence claim of
Section~\ref{sec:reindex} into an assertion a caller can make for itself.

\textbf{Retention as demotion.} Nothing above bounds growth, and an
agent memory accumulates low-value structure indefinitely. An importance score
over retrieval recency, retrieval frequency, degree centrality and belief-time
age marks the least valuable entities \emph{archived}: withheld from retrieval
and from community construction exactly as dormant entities are, reversible,
and preserved. Deletion is available but is neither the default nor silent. A
system whose claim is that history is kept cannot quietly discard it, and the
scoring requires telemetry to be enabled --- without it the recency and
frequency terms are undefined, and the call refuses rather than scoring on
structure alone.

\section{Retrieval and Synthesis}
\label{sec:retrieval}

\subsection{Modes}

Retrieval exposes six modes (Table~\ref{tab:modes}) selecting which evidence sources are consulted. An unknown mode raises rather than falling back to a default, since silently answering in a different mode than requested is indistinguishable from answering badly.

\begin{table}[t]
\centering
\small
\caption{Retrieval modes and the evidence each consults. \texttt{mix} is the default. Relations reach the prompt by three independent routes --- $k$-hop traversal from a matched entity, direct similarity search over relation embeddings, and lexical full-text search --- fused by reciprocal rank, as described in Section~\ref{sec:threechannel}. \texttt{bypass} performs no retrieval and exists so a caller can route a conversational turn through the same interface without a special case.}
\label{tab:modes}
\begin{tabular}{@{}lccccL{3.7cm}@{}}
\toprule
\textbf{Mode} & \textbf{Entities} & \textbf{Chunks} & \textbf{Relations} & \textbf{Communities} & \textbf{Intended question} \\
\midrule
\texttt{mix}    & \checkmark & \checkmark & $k$-hop $+$ similarity & --- & General; the default \\
\texttt{local}  & \checkmark & \checkmark & $k$-hop $+$ similarity & --- & About specific named things \\
\texttt{global} & ---        & ---        & ranked & \checkmark & Corpus-level themes \\
\texttt{hybrid} & \checkmark & \checkmark & $k$-hop $+$ sim.\ $+$ ranked & \checkmark & Spans both \\
\texttt{naive}  & ---        & \checkmark & ---   & --- & Plain vector RAG baseline \\
\texttt{bypass} & ---        & ---        & ---   & --- & No retrieval \\
\bottomrule
\end{tabular}
\end{table}

\subsection{Query conditioning and expansion}

A question is first decomposed into dual-level keywords \citep{guo2024lightrag}: low-level keywords naming concrete entities, high-level keywords naming themes. These are used asymmetrically. Low-level keywords are appended to the question before embedding the \emph{entity} query vector, sharpening entity search towards the named things; high-level keywords steer relation ranking in global mode. Keyword extraction is the one place a lexical fallback is acceptable, because a weak keyword list is never persisted and only widens retrieval --- unlike a weak extraction, which is written to the graph forever.

Retrieval then proceeds in three expansions. Vector search over \texttt{entities} and \texttt{documents} produces seeds, with dormant entities filtered out. Bounded traversal of up to $k$ hops from each matched entity, scoped to the caller's space, produces relations. Finally --- the payoff of the \texttt{doc\_mentions} edges --- chunks that \emph{mention} a matched entity are pulled in even if they did not match the query vector, ranked by how many of the matched entities they mention. A question can name an entity while the passage that explains it uses none of the question's wording, and that passage is frequently in a different document; without mention expansion it is unreachable.

Traversal depth defaults to $k=2$. One hop answers \emph{what is said about $X$}; it cannot answer \emph{how did $X$ come to cause $Y$}, because the edges carrying that chain lie between $X$'s neighbours and are never incident to $X$. Crucially, the temporal predicates of Section~\ref{sec:temporal} and any relation-type restriction are applied \emph{within} the walk rather than to its result. The distinction is not cosmetic: filtering afterwards still permits a path to travel \emph{through} a superseded or out-of-period edge to reach a vertex that then presents as current context. The offending edge vanishes from the output while the conclusion it enabled survives. Constraining each step means such paths are never walked. Because fan-out compounds --- three hops from one well-connected Boeing entity reaches over 25{,}000 edges --- \texttt{max\_relation\_edges} caps what any single matched entity may contribute.

Triples found by both the traversal and the global pass are de-duplicated on $(\text{subject}, \text{predicate}, \text{object})$ case-insensitively, then filtered by the temporal predicates of Section~\ref{sec:temporal}. In global mode without relation embeddings, relations are ranked by keyword overlap with corroboration weight as a tie-break, rather than returned in whatever order the rows were read.

Within the traversal channel, relations are ordered nearest-hop first and only then newest-asserted first. Ranking by assertion time alone is correct at one hop and quietly wrong beyond it, since assertion time across a corpus is close to arbitrary: a three-hop edge that happened to be indexed last would displace an adjacent one the moment a relation token budget truncated, which is the case whenever a caller sets one. Under hop-major ordering, two and three hops frequently synthesise identical answers, because truncation retains the same nearest relations either way --- depth costs more without degrading.

\subsection{Three-channel relation retrieval}
\label{sec:threechannel}

Traversal can only rank what it reached. A relation whose endpoints are generically named --- a charge, a programme, a line item the corpus never christens --- is not incident to any vertex the question matches, so no hop budget reaches it and no re-ranking recovers it. This is a property of candidate generation, not of ordering, and the two are separable experimentally: re-ranking the entire multi-hop candidate set by direct query--relation cosine similarity, a strictly better ranker over the same candidates, moved the on-topic share of retrieved relations from 67.8\% to 67.5\% --- that is, not at all.

Each distinct triple is therefore embedded at index time from its rendered surface form and stored on the edge, giving the edge table its own HNSW index; retrieval searches relations directly, in parallel with traversal. This is enabled by default from version 1.5.0. A third channel, added in 1.6.0, retrieves relations by PostgreSQL full-text search over the same rendered form \citep{robertson2009bm25}, behind a GIN index built on first use. It exists for the case neither of the others covers: an exact term --- an identifier, a statute, a part number --- that is lexically decisive and semantically unremarkable, and that a dense encoder therefore places near many irrelevant neighbours.

Merging three ranked lists is the substantive design question, and pooling by score is not available: the candidate sets are ranked by incommensurable quantities, and sorting a pooled set by similarity awards \emph{every} slot to the similarity channel, because that channel is ranked by precisely the quantity being sorted on. We verified this --- a pooled ranking reproduced the similarity channel's output exactly on all four evaluation questions.

Two merge strategies are implemented. The original is a fixed-share interleave to a quota $q \in [0,1]$ giving the similarity channel its share of the available relation slots. The default since 1.6.0 is reciprocal rank fusion (RRF) \citep{cormack2009rrf}, which scores each candidate by $\sum_c 1/(k + r_c)$ over the channels $c$ that returned it at rank $r_c$. RRF needs no share constant, and --- the reason it became the default --- it lets a third channel contribute without re-tuning the first two. Under a quota the lexical channel changed nothing at all, because a quota interleaves traversal against one seeded list and appended lexical results sit behind entries that a truncating budget removes --- which was the default when this was measured.

Measured through the shipped query path on the Boeing corpus, mean on-topic share of the 25 relations reaching the prompt is 92\% under RRF against 81\% under the quota, with RRF ahead on three of four questions. It loses the fourth, and the loss is interpretable: that question names a well-connected entity which traversal already covers completely, so fusing in two further channels can only dilute a set that was already correct. A corpus of such questions is the case for retaining the quota, which is why both ship.

One methodological point governs every figure above: all of them are measured through \texttt{query\_data} itself rather than through an offline reconstruction of the channels. The distinction is worth more than twenty points here --- an offline harness that rebuilds the merge by hand does not reproduce the entity seeding and hop-ordered filtering the shipped path applies, and understates the quota baseline accordingly. Relative comparisons drawn from such a harness remain informative, since both arms share its simplifications, but absolute shares should be taken from the retrieval path that runs in production.

\subsection{Diversification and reranking}
\label{sec:diversification}

Fusion decides what enters the candidate set; it says nothing about redundancy within it. Three further mechanisms, all adapted from \citet{rasmussen2025zep} and all off by default, address gaps the fused ranking leaves. Each is off because each costs something --- two cost ranking fidelity, one costs an LLM call per relation --- and because the evidence for them is a 20-instance paired sample rather than the full benchmark (Section~\ref{sec:paired}).

\textbf{Maximal marginal relevance (MMR)} \citep{carbonell1998mmr} trades relevance for coverage, keeping a candidate only if it adds something the selected ones do not. It matters here specifically because the channels correlate: traversal and relation-embedding search routinely return the same fact worded differently, and at a $\text{top-}k$ of eight a restatement costs a slot outright. The trade-off parameter is $\lambda = 0.7$ by default; above roughly $0.8$ the penalty rarely changes the order, below roughly $0.4$ it starts promoting unrelated relations over the ones that answer the question.

\textbf{Node-distance reranking} ranks by graph distance from the entities the question matched. The traversal channel carries true hop counts, but the relation-embedding and lexical channels report every result as one hop, because they never walked to what they found. Resolving a real distance for those stops a fact three hops from anything the question mentioned outranking one sitting directly on it --- an artefact introduced by the very channels that fixed the recall problem.

\textbf{LLM contradiction detection} complements declarative supersession. Exclusive predicate groups fire only on predicate pairs declared in advance between the same ordered pair, so they cannot see \texttt{lives\_in Paris} becoming \texttt{lives\_in Berlin}, where the predicate is identical and the object is what changed. The model is asked, but only about the residue the declarative pass did not resolve: the deterministic path runs first and the model never sees what it already handled. This is a write-path mechanism, unlike the other two.

\subsection{Synthesis}
\label{sec:synthesis}

The prompt is assembled from four blocks --- community reports, document passages, entities, relations --- with community reports leading when present because they carry corpus-level structure that no single passage holds. Each block may be capped at its own share of a token budget, but no cap is applied by default: everything retrieved reaches the reader unless a caller sets one for cost or for a small context window. Section~\ref{sec:context} is why, and measures what the previous default cost. Chunks are numbered and accompanied by a reference list so citations resolve to documents.

Denied relations are rendered explicitly, as \texttt{(A) --[NOT worked\_with]--> (B)}. This is the read-side half of the negation design: stored as a positive predicate with a flag, a denial rendered naively into the prompt would read to the model as an assertion that the relation holds, which is worse than not having extracted it. Retrieved relations carry their validity periods into the prompt, rendered as \texttt{[from 2023-05-10]} or \texttt{[valid 2022-01-01 to 2022-03-31]} beside the predicate. This is what lets an answer describe a trajectory rather than a timeless state, and it is what makes ordering and duration questions answerable at all: a model asked which of two events came first can only answer from dates it can see. Section~\ref{sec:validity} ablates it.

\section{Evaluation}
\label{sec:evaluation}

\subsection{Setup}

We evaluate on two public benchmarks and three corpora of deliberately different register. The benchmarks supply scored answer quality against reference answers; the corpora supply the graph-structural and retrieval measurements that explain it.

\begin{itemize}
\item \textbf{LongMemEval} \citep{wu2024longmemeval} --- 500 questions over long multi-session chat histories, in six categories, each instance carrying its own conversation set. Every instance is indexed from scratch and answered through the ordinary query path, and answers are scored against reference answers rather than inspected.
\item \textbf{ECT-QA} \citep{ectqa} --- time-sensitive question answering over earnings call transcripts, typed by temporal shape. We index six companies --- five in information technology and one retailer --- at sixteen consecutive quarters each, 96 transcripts covering 2020 to 2024, one space per company plus a combined space the cross-company questions query. It is the harder of the two by construction: a company restates the same metrics every quarter, so a metric such as free cash flow has sixteen values and the question names which one it wants.
\end{itemize}

The three corpora are:

\begin{itemize}
\item \textbf{Wikipedia} --- four articles (\emph{Ada Lovelace}, \emph{Charles Babbage}, \emph{Analytical engine}, \emph{Difference engine}), ${\sim}127$k characters, chosen because they share many entities, which makes them a test of cross-document resolution rather than of extraction alone.
\item \textbf{Dumas} --- the d'Artagnan trilogy, ${\sim}645$k characters: three novels following the same characters across roughly forty years, during which alliances genuinely reverse. Indexed in publication order.
\item \textbf{Boeing} --- five Boeing 10-K filings, Management's Discussion and Analysis only, 586{,}775 characters, spanning FY2006, FY2012, FY2018, FY2020 and FY2024. Those five points were chosen to straddle four operational cycles, so that relationships genuinely invert rather than being sampled evenly.
\end{itemize}

Embeddings are \texttt{text-embedding-3-small} (1536-d) throughout, chunking is 2000 characters with 200 of overlap, and each run indexes into its own PostgreSQL schema, recreated from scratch. LightRAG comparisons hold the extraction model, embedding model, gleaning depth and chunk sizing identical across both systems. The harness is the \texttt{evaluation/} directory of the reference implementation; exact commands are in \texttt{evaluation\_notes.md}.

LongMemEval graphs are extracted by \texttt{gemini-3.7-flash} and answered by \texttt{gemini-3.6-flash} in the headline configuration --- the split is deliberate, since Section~\ref{sec:readers} varies the answering model over fixed graphs; ECT-QA uses \texttt{gemini-3.6-flash} for both. All runs use \texttt{gemini-embedding-001} for embeddings, with the configuration frozen before the reported run; the corpus comparisons use the models named per table. On the corpora we report graph-structural measurements plus retrieval behaviour rather than scored answer quality, for the reason given in Section~\ref{sec:limitations}: a defensible cross-system answer-quality comparison requires the extraction model to be pinned across both systems, which a published baseline number never is. The benchmark is where scored answer quality is reported, and it carries its own comparability caveats, stated with it.

\subsection{End-to-end benchmark: LongMemEval}
\label{sec:longmemeval}

Table~\ref{tab:longmemeval} reports the full 500-question benchmark against the published Zep figures \citep{rasmussen2025zep} and a full-context baseline that places the entire conversation history in the prompt.

\begin{table}[t]
\centering
\small
\caption{LongMemEval, full benchmark. \texttt{post-graph-rag} drives \texttt{gemini-3.6-flash} with temporal grounding active (Section~\ref{sec:validity}), answers graded by a three-model majority panel; Zep figures are as published, driving \texttt{gpt-4o} and \texttt{gpt-4o-mini}. Best value per row in bold. One instance of 500 is excluded --- a session both extraction prompts refused --- and is reported here rather than absorbed, so the denominator is 499.}
\label{tab:longmemeval}
\begin{tabular}{@{}L{4.6cm}rrrr@{}}
\toprule
\textbf{Question type} & $n$ & \textbf{post-graph-rag} & \textbf{Zep} & \textbf{Zep} \\
& & \textbf{(flash)} & \textbf{(gpt-4o)} & \textbf{(4o-mini)} \\
\midrule
single-session-user & 70 & \textbf{95.7\%} & 92.9\% & 81.4\% \\
single-session-assistant & 55 & \textbf{100.0\%} & 80.4\% & 81.8\% \\
knowledge-update & 78 & \textbf{94.9\%} & 83.3\% & 76.9\% \\
multi-session & 133 & \textbf{90.2\%} & 57.9\% & 40.6\% \\
temporal-reasoning & 133 & \textbf{96.2\%} & 62.4\% & 36.5\% \\
single-session-preference & 30 & \textbf{83.3\%} & 56.7\% & 30.0\% \\
\midrule
\textbf{overall} & 499 & \textbf{94.0\%} & 71.2\% & 63.8\% \\
\bottomrule
\end{tabular}
\end{table}

\begin{figure}[t]
  \centering
  \includegraphics[width=0.98\linewidth]{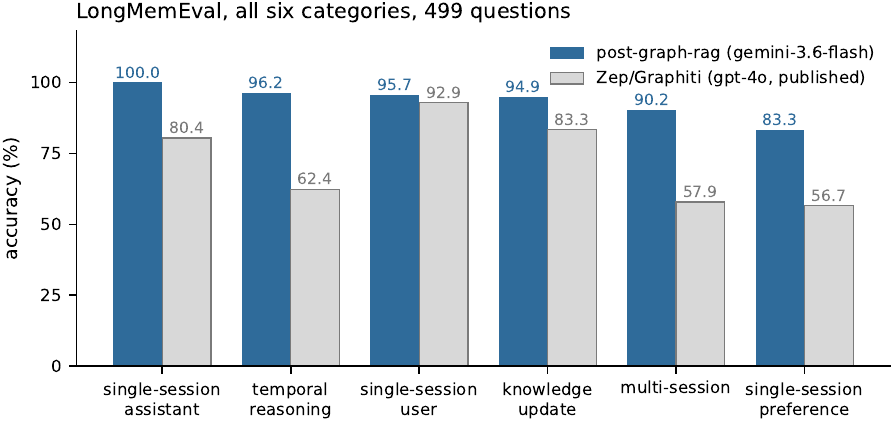}
  \caption{LongMemEval by category. The two widest margins fall on temporal reasoning and multi-session synthesis, the categories a bi-temporal graph exists to serve; a result whose largest gains had landed on single-session lookup would have warranted more suspicion. Zep figures are as published and carry the comparability caveats stated in the text.}
  \label{fig:longmemeval}
\end{figure}

A flash-class model leads the strongest published configuration on \emph{all six} categories and by 22.8 points overall, and the full-context baseline by 33.8. Figure~\ref{fig:longmemeval} shows the per-category comparison.

\textbf{These figures supersede those of the previous version of this paper, and the reason is a defect rather than a change of method.} Section~\ref{sec:context} describes it: context assembly discarded passages the retriever had already found, so every number reported in v2 was measured on an engine that withheld evidence from its own reader. The headline moves from 85.8\% to 94.0\% on the same corpus, same questions, same panel, same frozen configuration. We report both, and the ablation that separates them, because a revision that quietly restates a headline is worth less than one that explains it.

Absolute figures on a judged benchmark are protocol-dependent, and we make no claim here about how far this one would move under a different panel: the runs that would have supported such a claim were measured on the engine before the repair of Section~\ref{sec:context} and are not restated. What does not depend on the protocol is the paired ablation of Section~\ref{sec:validity}, which holds the graph, the questions and the judges fixed and varies one flag. The two widest margins fall where a bi-temporal graph is supposed to earn its keep: \textbf{temporal-reasoning} at $+33.8$ and \textbf{multi-session} at $+32.3$, the latter being the category Zep themselves score lowest on with \texttt{gpt-4o}. A result whose largest gains had landed on single-session lookup would have warranted more suspicion than this one does.

The comparison is not perfectly controlled, and the imbalances run in both directions. The models differ in both cost class and generation: their \texttt{gpt-4o} was the frontier tier of its day where ours are the low-cost flash tier of theirs, but their figures predate our models by roughly two years, so the direction of the model gap cannot be signed. They judge with a single \texttt{gpt-4o} where we use a three-model majority panel of gemma-4-26b, Llama-3.3-70B and gpt-oss-120b, with the answering model excluded from its own jury --- which is a difference of protocol rather than of strength. We report the comparison as informative about the tier, not as a controlled head-to-head --- the same standard Section~\ref{sec:models} applies to everyone else.

\textbf{The two hardest categories are the two that turn on time.} Knowledge-update requires distinguishing a fact from the fact it replaced; temporal-reasoning requires ordering events or measuring an interval between them. Both are won on the read side, by carrying validity through to synthesis: Section~\ref{sec:validity} ablates that mechanism at $+38.4$ points on temporal-reasoning and $+16.3$ on knowledge-update. That ablation was measured before the repair of Section~\ref{sec:context}, so its absolute levels belong to the older engine; what it establishes is that both categories are won on the read side, which is where the headline figures above put them.

\subsection{Context assembly: how much of what was retrieved reaches the reader}
\label{sec:context}

Retrieval quality is usually measured as though everything retrieved arrives at
the model. Between the two sits an assembler that fits passages into a token
budget, and a defect there is invisible to every retrieval metric: recall is
computed over what was found, not over what survived the walk to the prompt.

The engine used for the previous version of this paper carried a 4000-token
context budget, divided by channel --- half to documents, the remainder shared
between entities, relations and community reports. Passages were selected
greedily until one exceeded the remaining budget, at which point selection
stopped. A passage larger than its channel's share therefore terminated
selection before anything had been selected, and the channel contributed
\emph{nothing}. This is not a rare corner: a LongMemEval haystack session runs
to some fourteen thousand characters against a two-thousand-token document
share, and a section of an SEC filing is larger still. The model answered
``no information available'' while holding, unread, the passages that contained
the answer.

Two changes follow. Truncation clips the passage that overflows instead of
discarding it, so a budget degrades gracefully rather than to zero. And the
budgets themselves default to unlimited: everything retrieved reaches the
reader unless a caller sets a cap for cost or for a small context window.
Withholding retrieved evidence is no longer the default behaviour.

\begin{figure}[t]
  \centering
  \includegraphics[width=0.98\linewidth]{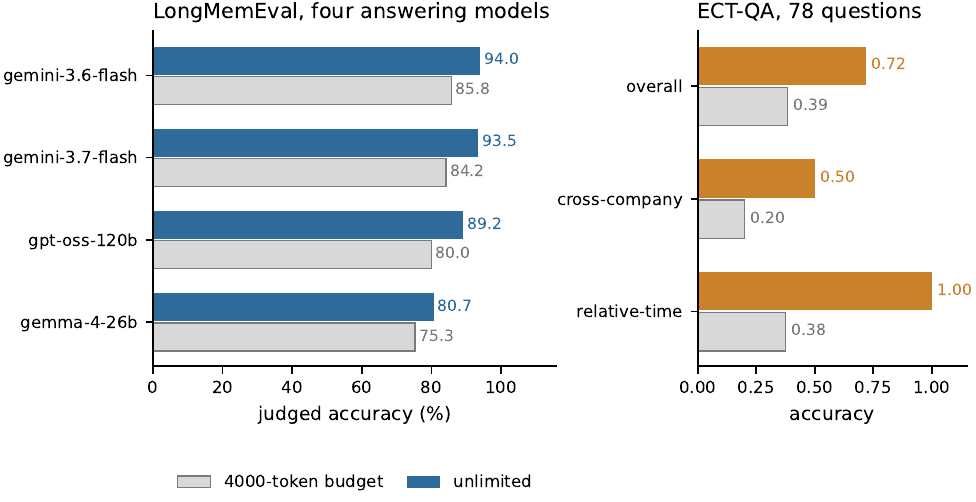}
  \caption{Context assembly, ablated on fixed graphs. Same corpora, questions, judges and configuration; only the assembler differs. Every answering model gains, which is what distinguishes a repair from a tuning choice --- a change that helped one model and hurt another would not support the same reading. The ECT-QA panel is the same realm and the same 78 questions under the current scorer, with only the assembler differing; there the loss was near-total, since retrieval fetches 48 chunks of filing prose against the same document budget.}
  \label{fig:context}
\end{figure}

Figure~\ref{fig:context} ablates it. The effect is large and it is uniform in
sign: every answering model improves, by five to nine points.

 On ECT-QA the same ablation is larger still. Retrieval there fetches
forty-eight chunks of filing prose against the same two-thousand-token share, so
almost nothing survived: 0.385 against 0.718 overall on the same graph and the
same questions, with relative-time moving from 0.375 to 1.000 and cross-company
--- which needs figures from several companies at once, and is therefore the most
dependent on many passages surviving --- from 0.200 to 0.500.

The finding worth carrying beyond this system is methodological. A cross-company
deficit on ECT-QA had survived three separate falsification attempts ---
retrieval coverage, refusal wording, and validity windows --- because all three
looked upstream of synthesis, where the loss was actually occurring. An
assembler that silently drops evidence is indistinguishable, from every metric
we had, from a retriever that never found it. Systems of this shape should
instrument the boundary: how many retrieved characters reach the prompt is a
number worth logging, and had it been logged the defect would have been visible
immediately rather than after three hypotheses had been tested and discarded.

\subsection{Temporal grounding in the prompt}
\label{sec:validity}

The temporal layer of Section~\ref{sec:temporal} extracts validity from prose, normalises it, promotes it from JSONB to a generated column and indexes it. Section~\ref{sec:synthesis} then renders it beside each relation. This section ablates that last step, because it is the one that converts stored temporal structure into answerable questions, and its contribution turns out to be the largest of any single mechanism we measure.

The questions it serves are the ones whose answer is an ordering or an interval --- which of two events came first, how long between them. Both endpoints may sit in the graph, correctly dated, and remain unusable if the dates are not in front of the model when it answers.

\begin{table}[t]
\centering
\small
\caption{Ablating temporal grounding in the prompt, LongMemEval, paired. Each instance is indexed once and four answering models read that identical graph; \texttt{render\_relation\_validity} is the only difference between arms. Columns are the mean over the four readers; ``better'' and ``worse'' count instances whose summed score across readers moved.}
\label{tab:validity}
\begin{tabular}{@{}L{4.9cm}rrrrrr@{}}
\toprule
\textbf{Question type} & $n$ & \textbf{off} & \textbf{on} & \textbf{delta} & \textbf{+} & \textbf{--} \\
\midrule
temporal-reasoning & 133 & 0.496 & \textbf{0.881} & \textbf{+0.384} & 69 & 9 \\
knowledge-update & 78 & 0.663 & \textbf{0.827} & \textbf{+0.163} & 30 & 7 \\
single-session-preference & 30 & 0.567 & 0.633 & $+0.067$ & 6 & 2 \\
multi-session & 133 & 0.641 & 0.673 & $+0.032$ & 22 & 23 \\
single-session-user & 70 & 0.939 & 0.961 & $+0.021$ & 3 & 2 \\
single-session-assistant & 55 & 0.868 & 0.877 & $+0.009$ & 2 & 2 \\
\midrule
\textbf{overall} & 499 & 0.668 & \textbf{0.813} & \textbf{+0.145} & & \\
\bottomrule
\end{tabular}
\end{table}

\textbf{The single-session categories are the control, and they behaved.} Any change that puts more text in the prompt risks buying a general uplift and attributing it to a mechanism. The two categories in which everything needed sits inside one conversation move by 2.1 and 0.9 points, and multi-session's 22-better-against-23-worse is noise. The prediction --- dates should move the categories that need dates and leave the rest alone --- was recorded before the run. On \texttt{gemini-3.6-flash} temporal-reasoning goes from 0.519 to 0.932 with 55 instances better and none worse.

\textbf{The result locates where temporal capability is won.} Section~\ref{sec:paired} reports two index-side candidates --- contradiction detection, and broadened extraction of dated actions --- neither of which moved the benchmark. Both aimed at putting \emph{more} temporal structure into the graph. The mechanism that pays is on the read side, and by a wide margin: what a temporally-aware system needs is not primarily more extracted dates but the discipline of carrying the dates it has through to the point of synthesis. We would encourage treating the render path as a first-class part of a temporal design rather than a formatting detail.

\textbf{The same change bought nothing on ECT-QA} (Section~\ref{sec:ectqa}), which is what makes the mechanism credible rather than a general improvement. That corpus states its quarter in the indexed text of every transcript, so the model was never short of period information, and rendering it moved one or two questions per arm.

\subsection{How the configuration was frozen}
\label{sec:frozen}

The reported figure is the frozen configuration described here plus the temporal grounding of Section~\ref{sec:validity}; this section records how the configuration itself was reached, because the discipline is where the number's credibility lives.

The configuration was frozen before the full run, after development against a 120-instance stratified sample on a single seed. That discipline is not ceremonial. The same code had already scored 75\% on a favourable 20-instance draw, with temporal-reasoning at 81\% against the 52.6\% it reaches on the full set. At these sample sizes the variance is not a rounding concern; it is the difference between ``beats the published baseline'' and ``does not''.

Development was as instructive for what it rejected. Four candidate improvements were tested on the development seed and two survived: an answer-side rule resolving conflicting records to the most recent statement, worth twenty points on knowledge-update, and wider retrieval at $\text{top-}k = 32$. A refinement distinguishing updates from accumulating preferences regressed and was dropped. Query decomposition for two-event questions --- mechanistically the best-motivated of the four --- cost 6.7 points and was dropped from the configuration while remaining in the library, off by default. A held-out seed then showed the development results themselves swinging by 15 to 35 points per category at 12 to 31 instances per type, which is why the reported figure comes from the full set and from nothing smaller.

\subsection{Contribution of each mechanism}
\label{sec:ablation}

\begin{figure}[t]
  \centering
  \includegraphics[width=\linewidth]{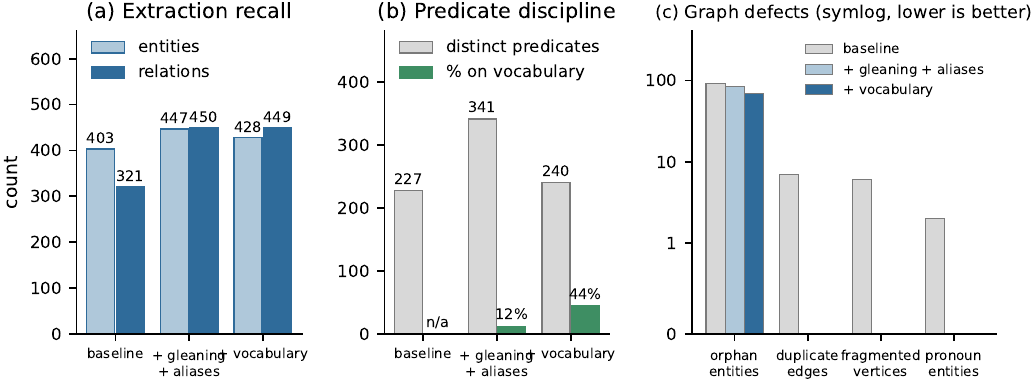}
  \caption{Contribution of each mechanism on the Wikipedia corpus, extraction model held fixed. (a) Gleaning raises relation recall by 40\% (321 $\to$ 450); the vocabulary costs almost none of it back. (b) Free-form extraction produces a predicate per relation --- 341 distinct predicates over 450 relations --- which the vocabulary reduces to 240 while raising the share of relations carrying a queryable predicate from 12\% to 44\%. (c) Structural defects, symlog scale. Duplicate edges, short-form fragmentation and pronoun vertices go to zero once corroboration weighting, alias resolution and the pronominal gate are active; orphan entities fall further under the vocabulary because snapping attaches relations that would otherwise have been dropped as unmatched.}
  \label{fig:ablation}
\end{figure}

Table~\ref{tab:ablation} and Figure~\ref{fig:ablation} isolate the mechanisms on the Wikipedia corpus, holding the extraction model fixed so the columns reflect the features rather than the model.

\begin{table}[t]
\centering
\small
\caption{Effect of each mechanism, Llama-3.3-70B throughout, 40 chunks. ``Vocabulary predicate'' is the share of stored relations whose predicate is in the 35-term \texttt{biography} vocabulary. ``Fragmented vertices'' counts entities split across short and long surface forms; ``pronoun entities'' counts vertices whose name is a pronoun or relative reference. Best value per row in bold. The middle arm indexed 39 of 40 chunks --- one chunk's gleaning pass returned nothing usable and the chunk was skipped --- so its counts come from marginally less text than the arms either side, which understates gleaning rather than flattering it.}
\label{tab:ablation}
\begin{tabular}{@{}L{6.0cm}rrr@{}}
\toprule
& \textbf{baseline} & \textbf{$+$ gleaning \& aliases} & \textbf{$+$ vocabulary} \\
\midrule
Chunks indexed & \textbf{40}/40 & 39/40 & \textbf{40}/40 \\
Entities & 403 & 447 & 428 \\
Relations & 321 & 450 & 449 \\
Distinct predicates & 227 & 341 & \textbf{240} \\
Relations on a vocabulary predicate & --- & 12\% & \textbf{44\%} \\
Orphan entities (no relations) & 92 & 85 & \textbf{69} \\
Duplicate edge rows & 7 & \textbf{0} & \textbf{0} \\
Fragmented short-form vertices & 6 & \textbf{0} & \textbf{0} \\
Pronoun entities & 2 & \textbf{0} & \textbf{0} \\
\bottomrule
\end{tabular}
\end{table}

Three effects separate cleanly. \textbf{Gleaning buys recall}: relations rise 40\% and entities 11\%, at one extra LLM call per chunk. \textbf{Alias resolution eliminates fragmentation}: six split vertices become zero --- \texttt{Charles Babbage} absorbs \texttt{Babbage} and \texttt{Mr.\ Babbage}, \texttt{Ada Lovelace} absorbs six surface forms --- and corroboration weighting takes duplicate edge rows from seven to zero. \textbf{The vocabulary buys queryability}: it does not change how many relations exist, but it changes whether they can be filtered by type. Without it the head of the predicate distribution is flat, the commonest predicate appearing six times; with it the head is real --- \texttt{designed}~(20), \texttt{worked\_with}~(18), \texttt{built}~(17), \texttt{wrote}~(14), Figure~\ref{fig:partition}(b).

Cross-document linkage holds throughout: \texttt{Charles Babbage} and \texttt{Analytical Engine} are each reached from all three articles that mention them, and a question about the two engines retrieves chunks from both engine articles.

\subsection{Sensitivity to the extraction model}
\label{sec:models}

\begin{figure}[t]
  \centering
  \includegraphics[width=\linewidth]{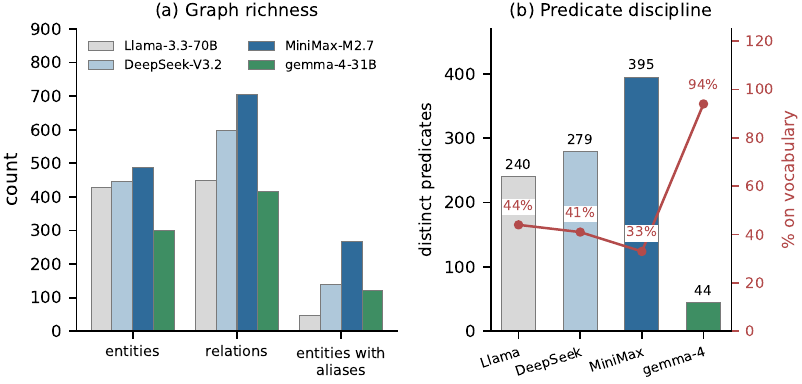}
  \caption{Four extraction models, everything else held fixed. (a) MiniMax-M2.7 builds the richest graph, with 268 entities carrying aliases --- $5.7\times$ Llama's --- which is what drives its low orphan count. (b) The two axes move in opposition: gemma-4-31B produces only 44 distinct predicates at 94\% vocabulary adherence, MiniMax 395 at 33\%.}
  \label{fig:models}
\end{figure}

Holding corpus, settings and clustering fixed and varying only the extraction model across four models produces Table~\ref{tab:models} and Figure~\ref{fig:models}.

\begin{table}[t]
\centering
\small
\caption{Effect of the extraction model. Identical corpus, identical settings (gleaning 1, \texttt{biography} vocabulary, Leiden at $\gamma=2.0$), 40/40 chunks each; only the primary model differs. Best value per row in bold (fewer orphans and fewer single-use predicates are better).}
\label{tab:models}
\begin{tabular}{@{}L{4.15cm}rrrr@{}}
\toprule
& \textbf{Llama-3.3-70B} & \textbf{DeepSeek-V3.2} & \textbf{MiniMax-M2.7} & \textbf{gemma-4-31B} \\
\midrule
Entities & 428 & 447 & \textbf{488} & 299 \\
Relations & 449 & 599 & \textbf{705} & 417 \\
Entities carrying aliases & 47 & 139 & \textbf{268} & 120 \\
Negated relations captured & 0 & \textbf{21} & 7 & 14 \\
Corroborated (weight $>1$) & 16 & 35 & 28 & \textbf{44} \\
Orphan entities & 69 & 41 & \textbf{26} & 31 \\
Distinct predicates & 240 & 279 & 395 & \textbf{44} \\
Single-use predicates & 74\% & 67\% & 74\% & \textbf{25\%} \\
Relations on vocabulary & 44\% & 41\% & 33\% & \textbf{94\%} \\
Community \texttt{rating} spread & 1 value & 3 (7--9) & 3 (7--9) & \textbf{6 (4--10)} \\
\bottomrule
\end{tabular}
\end{table}

\textbf{There is no single winner; the models split into two useful shapes.} MiniMax-M2.7 builds the \emph{richest} graph --- most entities, most relations, and $5.7\times$ Llama's alias emission, which is what drives its orphan count down to 26. Its community reports are also the most insightful, producing findings such as ``ENIAC designers worked independently without knowledge of Babbage's analytical engine'' rather than restatements of the cluster label. gemma-4-31B builds the most \emph{queryable} graph --- 44 distinct predicates at 94\% vocabulary adherence, against 33--44\% for everything else, with top predicates carrying real mass (\texttt{located\_in} 38, \texttt{worked\_with} 36, \texttt{studied} 34) where MiniMax's heaviest appears only 19 times. It also produced the most corroborated edges and by far the best community rating discrimination. The cost is recall: it stores 40\% fewer relations, because it snaps aggressively onto the vocabulary rather than preserving distinctions. Choose MiniMax to query by similarity and traversal, gemma to query by relation type. Llama-3.3-70B is best on none of these measures.

The methodological point is the one that generalises: \textbf{features depending on the model doing something subtle fail silently on weak models.} Aliases only merge if the model emits them, and Llama never once used the \texttt{negated} field on this corpus --- silently collapsing ``X worked with Y'' and ``X never met Y'' into the same edge. Any comparison between Graph RAG systems that does not hold the extraction model fixed is measuring the model, not the system.

\subsection{Comparison with LightRAG on encyclopedic prose}
\label{sec:lightrag}

\begin{figure}[t]
  \centering
  \includegraphics[width=0.86\linewidth]{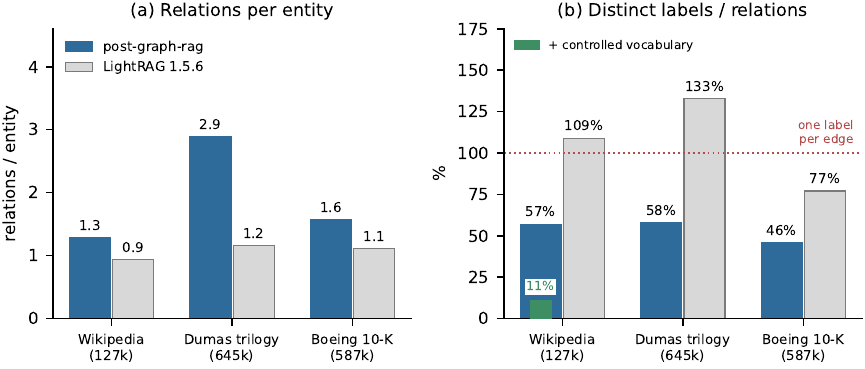}
  \caption{Head to head against LightRAG across all three corpora, identical extraction and embedding models throughout. (a) \texttt{post-graph-rag} extracts the denser graph on every corpus, by up to $2.4\times$ on relations per entity. (b) Distinct edge labels as a share of relation count --- lower is more queryable. LightRAG exceeds 100\% on two of three corpora, meaning more unique labels than edges; \texttt{post-graph-rag} stays at 46--58\% unconstrained; the 11\% under a controlled vocabulary (green) is \texttt{gemma-4-31B}, whose adherence is the highest we measured, rather than the model used for the head-to-head. Indexing wall-clock is not plotted: the two systems write to different substrates under different durability guarantees, so a throughput ratio would not be attributable to the extraction pipeline.}
  \label{fig:lightrag}
\end{figure}

Both libraries indexed the same four articles with the same extraction model (MiniMax-M2.7), the same embedding model, one gleaning pass, and equivalent chunk sizing (Table~\ref{tab:lightrag}, Figure~\ref{fig:lightrag}).

\begin{table}[t]
\centering
\small
\caption{Head to head on the Wikipedia corpus. Identical extraction model, embedding model, gleaning depth and chunk sizing. Character counts differ because \texttt{post-graph-rag}'s chunker discards section headings and short fragments, so the per-10k rows are the comparable density measures. Better value per row in bold. Indexing wall-clock is not tabulated; see the note on comparability below. \textsuperscript{$\dagger$}The 11\% is \texttt{gemma-4-31B} under the \texttt{biography} preset (44 labels over 417 relations), not the \texttt{MiniMax-M2.7} run the rest of this column reports: vocabulary adherence is a joint property of the preset and the model, and the same preset under \texttt{Llama-3.3-70B} reaches only 53\% (Table~\ref{tab:ablation}). It is included because it bounds what the mechanism can do, not because it is this column's configuration.}
\label{tab:lightrag}
\begin{tabular}{@{}L{5.6cm}rr@{}}
\toprule
& \textbf{post-graph-rag} & \textbf{LightRAG 1.5.6} \\
\midrule
Entities & \textbf{523} & 450 \\
Relations & \textbf{671} & 421 \\
Entities per 10k characters & \textbf{78.8} & 59.4 \\
Relations per 10k characters & \textbf{101.1} & 55.6 \\
\midrule
Distinct edge labels $\div$ relations & \textbf{57\%} & 109\% \\
\quad with a controlled vocabulary\textsuperscript{$\dagger$} & \textbf{11\%} & not supported \\
\midrule
Query latency, \texttt{mix} & \textbf{8.2 s} & 10.1 s \\
Query latency, \texttt{global} & \textbf{3.1 s} & 6.1 s \\
\bottomrule
\end{tabular}
\end{table}

\paragraph{Graph density.} \texttt{post-graph-rag} extracts roughly \textbf{33\% more entities and 82\% more relations per unit of text}. The gap is widest on relations, which is where gleaning pays off.

\paragraph{Edge labelling --- the sharpest structural difference.} LightRAG's edge \texttt{keywords} are free text: 460 distinct labels across 421 relations, i.e.\ \emph{more than one unique label per edge on average}, with entries such as ``social contact'' and ``claimed influence''. These describe an edge in prose but cannot be queried as a relation type; \texttt{WHERE relation\_type = 'worked\_with'} has no meaning against them. \texttt{post-graph-rag} normalises predicates and optionally snaps them onto a controlled vocabulary, sitting at 57\% unconstrained here. The best we measure under a preset is \textbf{11\%} --- 44 labels over 417 relations, with a real head to the distribution --- but that is \texttt{gemma-4-31B}, and the same preset under \texttt{Llama-3.3-70B} reaches 53\%. Adherence is a joint property of the vocabulary and the model that snaps onto it, which is Section~\ref{sec:models}'s point arriving in a different place. What holds regardless of model is that a normalised predicate can be filtered on and a free-text keyword cannot; if you intend to traverse or filter by relation type, that difference is decisive.

\paragraph{Retrieval.} Answer quality is comparable: both produce well-structured, accurate answers to factual and comparative questions. \texttt{post-graph-rag} is \textbf{faster at query time on both modes} --- notably $2\times$ on \texttt{global} --- and adds inline citations resolving to document chunks, which LightRAG does not. Corpus-level questions are served by the importance-weighted community ranking of Section~\ref{sec:communities}, under which both systems answer ``what are the main themes?'' centrally.

\paragraph{Why we do not report an indexing-throughput comparison.} Wall-clock indexing time is the one measurement we deliberately leave out throughout, and the cross-system case is the clearest reason why: the two systems are not performing the same operation. \texttt{post-graph-rag} commits every entity and relation to a transactional database through a connection, resolving each entity against a uniqueness index as it goes; LightRAG's default backend mutates an in-process NetworkX object backed by a file. Those are different durability and consistency guarantees, and they impose different concurrency limits --- ordered graph writes within a document for one, unrestricted parallelism across documents for the other (Section~\ref{sec:concurrency}).

A wall-clock ratio between them therefore measures the storage architecture at least as much as the extraction pipeline, and it would change under a different LightRAG backend without any code in either system changing. We record the raw figures in \texttt{evaluation\_notes.md} for completeness rather than presenting them as a comparison. What can be said without confounding is that the ordered-write design is a one-off index-time cost, that concurrent indexing reduced it by roughly $3.4\times$ on the Wikipedia corpus, and that at \emph{query} time --- where both systems read from their respective stores under the same conditions --- \texttt{post-graph-rag} is the faster of the two on both modes.

\paragraph{Operational differences.} \texttt{post-graph-rag} persists to PostgreSQL with realm/space multi-tenancy, audit tables and append-only history; LightRAG defaults to file-backed NetworkX plus nano-vectordb, with pluggable backends available. On a router that exhausted provider credits mid-run, LightRAG's first attempt failed three of four documents and the comparison only completed after retry and failover were added to the harness --- \texttt{post-graph-rag} has that built in, deadline-bounded so a sustained outage surfaces quickly.

\subsection{Corpora whose facts change}
\label{sec:temporal-eval}

\begin{figure}[t]
  \centering
  \includegraphics[width=\linewidth]{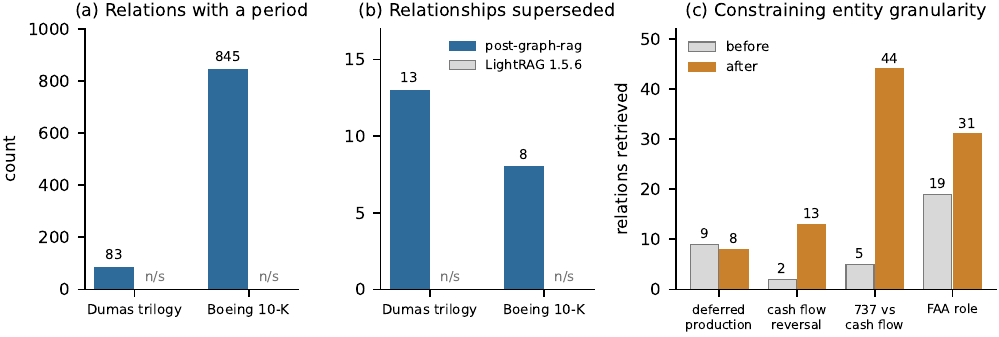}
  \caption{Temporal behaviour. (a,b) Relations carrying an extracted validity period, and relationships closed by a later document. LightRAG models neither, so \texttt{n/s} marks a structural absence rather than a low score. (c) Relations retrieved per question on the Boeing corpus before and after constraining entity granularity in the extraction prompt (Section~\ref{sec:granularity}). The two questions that started with the least evidence gain the most, up to $9\times$; the first question was already well served and is unchanged.}
  \label{fig:temporal}
\end{figure}

Encyclopedic prose does not exercise the temporal layer, because its facts do not move. The two corpora below do (Table~\ref{tab:temporal}, Figure~\ref{fig:temporal}).

\begin{table}[t]
\centering
\small
\caption{Temporal corpora, head to head. Same extraction model, embedding model and chunk sizing throughout; documents indexed in publication order (Dumas) and fiscal-year order (Boeing). Better value per row in bold. ``not modelled'' marks a capability the baseline does not have; ``n/r'' marks a figure we did not record for that run.}
\label{tab:temporal}
\begin{tabular}{@{}L{4.3cm}rrrr@{}}
\toprule
& \multicolumn{2}{c}{\textbf{Dumas trilogy} (645k)} & \multicolumn{2}{c}{\textbf{Boeing 10-K} (587k)} \\
\cmidrule(lr){2-3}\cmidrule(lr){4-5}
& \textbf{ours} & \textbf{LightRAG} & \textbf{ours} & \textbf{LightRAG} \\
\midrule
Entities & \textbf{1{,}377} & 1{,}246 & \textbf{3{,}078} & 2{,}832 \\
Relations & \textbf{3{,}976} & 1{,}447 & \textbf{4{,}830} & 3{,}147 \\
Relations per entity & \textbf{2.9} & 1.2 & \textbf{1.6} & 1.1 \\
Distinct labels $\div$ relations & \textbf{58\%} & 133\% & \textbf{46\%} & 77\% \\
Entities carrying aliases & \textbf{729} & not modelled & n/r & not modelled \\
Relations with stated validity & \textbf{83} & not modelled & \textbf{845} & not modelled \\
Relationships superseded & \textbf{13} & \textbf{0} & \textbf{8} & \textbf{0} \\
\bottomrule
\end{tabular}
\end{table}

\paragraph{Supersession fires on real prose.} On the Dumas trilogy, thirteen relationships were closed by a later book, including several the trilogy is built on: an \texttt{ally\_of} edge between d'Artagnan and Aramis closed once a later volume recasts them as opponents, and Rochefort's \texttt{friend\_of} link to the Cardinal closed as his allegiance shifts. All resolved from publication order alone --- the extractor supplied no dates for those pairs. On the Boeing corpus, eight relationships were superseded across the fiscal-year sequence.

\textbf{Density is the precondition.} An earlier run sampling ${\sim}5\%$ of each novel fired \emph{zero} supersessions, because no entity pair was characterised twice. Sampling evenly but sparsely does not help; the same pair must actually recur, which is a direct argument for the denser extraction of Figure~\ref{fig:lightrag}(a).

\paragraph{Edge labelling inverts further.} On narrative prose LightRAG reaches 133\% --- on average every edge carries its own unique label. Long-form fiction produces far more varied phrasing than encyclopedic text, so the gap between free-text keywords and normalised predicates widens rather than narrows with corpus difficulty.

\paragraph{As-of retrieval behaves as designed.} On Boeing, relation counts grow $22 \to 24 \to 25$ across $2006 \to 2024$, and the edge
\begin{quote}\ttfamily\small
(777X deferred production costs) --[reduced\_by]--> (777X program) \quad [2020-12-31..open]
\end{quote}
surfaces only at \texttt{as\_of=2024}. On the headline test --- \emph{trace how deferred production costs evolve from a minor line item to the central driver of cash burn} --- both systems answer, but differently: LightRAG returns a correct and \emph{timeless} accounting definition, while \texttt{post-graph-rag} returns the trajectory, because the retrieved relations carry periods.

\paragraph{Entity resolution on honorific-heavy fiction.} 729 entities carry aliases on the Dumas corpus, so the mechanism works at scale, but it is not exhaustive on this register: \texttt{M.\ d'Artagnan the younger}, \texttt{Monsieur d'Artagnan} and \texttt{D'Artagnan} remain distinct vertices, and \texttt{The Son of Henry IV} is a periphrasis for Louis XIII that nothing resolves. Fiction is harder here than Wikipedia.

\subsection{Financial prose is the adversarial case}
\label{sec:finance}

Every other corpus we tested is narrative, where entities are naturally canonical --- people, places, works. Filings are not, and they are what exposed the granularity problem of Section~\ref{sec:granularity}. Constraining entity granularity in the extraction prompt raised supersessions from 5 to 8, cut \texttt{\%boeing\%} vertices from 54 to 37, and lifted retrieval recall between $2\times$ and $9\times$ (Figure~\ref{fig:temporal}(c)). All four evaluation questions are answered substantively, and the 737 question produces an era-by-era trajectory rather than a summary.

The diagnosis is worth more than the fix, because it generalises. \texttt{query\_data} returns the retrieved entities, relations and chunks without synthesis, so a weak answer can be attributed to retrieval or to generation rather than guessed at. That is what identified the fault here: the two weakest answers were the two that retrieved fewest relations, which localised the problem to entity granularity at extraction time --- upstream of retrieval, and invisible from the answers alone. A system that only exposes final answers offers no equivalent handle, and the fix would have been sought in the retriever or the synthesis prompt, where it was not.

Two narrower guards proved necessary on this register and are now enforced in the extractor: entity names are length-capped, because a filing occasionally returns a whole table row as a name and the unique index backing entity resolution rejects it; and bare quantities are refused as vertices.

\textbf{Vocabulary adherence has a longer tail here.} The \texttt{finance} preset supplies 32 predicates and 8 exclusivity groups, and every group member was emitted --- \texttt{generates\_cash\_flow}~(80) against \texttt{consumes\_cash}~(8), \texttt{increases\_revenue}~(52) against \texttt{reduces\_revenue}~(56), \texttt{incurs\_charge}~(49) against \texttt{reports\_loss}~(47) --- so the mechanism works. But labels sit at 46\% of relation count rather than the ${\sim}11\%$ reached on biography. Filings discuss a far wider range of topics than an encyclopedia article, and a 32-term vocabulary covers the reversals without covering the tail. Expect to extend the vocabulary per domain rather than to inherit one.

\subsection{ECT-QA: time-sensitive questions, and what scoring them cost}
\label{sec:ectqa}

ECT-QA is the adversarial register of Section~\ref{sec:finance} turned into a scored benchmark. Sixty answerable questions across five temporal shapes, plus eight deliberately unanswerable ones, over a corpus in which every metric is restated sixteen times and only the date distinguishes the values.

That sequence produced \textbf{0.353 accuracy, mean numeric F1 0.451, with 6 of 60 questions refused} --- the figure this paper's previous version reported. It has since been superseded twice over, by the context-assembly repair of Section~\ref{sec:context} and by the two scoring corrections below, and the current figures are those of Table~\ref{tab:ectqa-types}. We keep the sequence because what arriving at any of these numbers required is the more useful result: three of the four changes in Table~\ref{tab:ectqa} were repairs to the \emph{evaluation}, not to the system, and the pattern did not stop there.

\begin{table}[t]
\centering
\small
\caption{ECT-QA, effect of scoring and prompt changes, \texttt{gemini-3.6-flash} throughout. The first two rows score \emph{identical} answers: only the metric differs. Accuracy counts a question correct at numeric F1 $\geq 0.60$; refusals are of the 60 answerable questions. \textbf{These are historical levels}, measured on a 68-question sample and on the engine before the repairs of Sections~\ref{sec:context} and~\ref{sec:ectqa}; the current figures are Table~\ref{tab:ectqa-types}. What the sequence establishes is the direction and cause of each step, which the repairs do not disturb.}
\label{tab:ectqa}
\begin{tabular}{@{}L{6.9cm}rrr@{}}
\toprule
& \textbf{accuracy} & \textbf{mean F1} & \textbf{refused} \\
\midrule
judge panel, original prompt & 0.162 & --- & 46/60 \\
numeric F1, same answers & 0.221 & 0.191 & 46/60 \\
\quad + extraction split, $\text{top-}k$ 48, partial answers & 0.324 & 0.448 & 4/60 \\
\quad + scope discipline & \textbf{0.353} & \textbf{0.451} & 6/60 \\
\bottomrule
\end{tabular}
\end{table}

\textbf{The judge panel scored worse than the system it was judging.} Asked for gross margin in each quarter of 2022, the system answered 33.9\%, 33.6\%, 31.7\% and 32.0\% against a gold of 33.9\%, 33.6\%, 31.7\% and 32.3\%: three exact matches and a fourth within 0.3 points. All three judges failed the answer outright. Rescoring identical answers by tolerance-based numeric F1 moved the run from 0.162 to 0.221 --- six points that were never a system failure. We think this is the more general hazard: an LLM judge asked for a verdict on a list of figures is being asked to do arithmetic it is not reliable at, and it fails in the direction that flatters the judge's own confidence.

\textbf{Cosine similarity was the obvious replacement and was rejected on measurement.} Gold answers are bare lists of figures while the system replies in prose, so whole-text embedding similarity is dominated by length. Correct answers averaged 0.643 and refusals 0.509 --- 0.13 apart, with no threshold separating them. Figure matching separated the same rows at 0.770 against 0.144. Cosine has since been removed from the harness altogether: a headline averaging it in was, in part, a similarity score rather than a figure measurement. The few questions whose gold names a company rather than a figure are read by a single judge instead.

\textbf{The answer prompt was manufacturing the refusals.} It ended \emph{``if the facts do not support an answer, say exactly: unanswerable''}, which the model read as demanding completeness: it refused 46 of 60 questions, several while quoting figures it had already retrieved. Under an F1 metric a partial answer earns partial credit, so the instruction was discarding points the metric would have awarded. Rewritten to invite partial answers, refusals fell to 4 of 60 and mean F1 rose from 0.191 to 0.448. Retrieval widened from $\text{top-}k$ 12 to 48 at the same time, since gold answers require a mean of 5.5 figures and as many as 32 --- typically one per quarter across sixteen quarters, which twelve chunks cannot cover however well they are ranked.

\textbf{A fourth change followed from watching what the third cost.} Inviting partial answers helped multi-period questions and hurt single-period ones: the model volunteered neighbouring quarters where exactly one figure was wanted, and precision charged it for them. Instructing it to answer the periods asked and no others --- and, where no period is named, the most recent the facts cover --- recovered that without giving back the gain.

\paragraph{A correction to our own metric.} The first numeric-F1 implementation counted every number in the answer text as a figure, and it cut both ways. A prose answer citing ``fiscal 2023-q2 [1]'' lost precision on chronology it never asserted as a value, so single-time questions that \emph{contained} their gold figures scored 0.38 to 0.59 against a 0.60 threshold and read as a category pinned at zero; meanwhile gold answers that state a year alongside each figure handed out free recall matches, inflating multi-time. Chronology and citation markers are now excluded from figure matching, and questions whose gold \emph{is} a period get period-token matching instead of falling through to cosine. Table~\ref{tab:ectqa-types} is lower than the version we first computed, and we state the correction rather than applying it silently.

\paragraph{A second correction, larger than the first: score what an answer asserts.} ``In which quarter was the margin highest'' is answered by one quarter, but answering it \emph{well} means naming the window searched and the runners-up compared. Period matching then counts the model's own justification against it, because its precision denominator holds every quarter the prose mentions while the gold holds one. Thirteen answers that named the correct quarter scored between 0.25 and 0.50 and were marked wrong. Classifying all 46 failures of that run: sixteen contained the gold period outright, four contained every gold figure, eleven were genuine partial answers, ten were declines, and five were real misses --- twenty of forty-six ``wrong'' answers held the full gold answer.

The repair is to extract the asserted answer before scoring it, which is what the surrounding literature does and this harness was not doing. FinQA and ConvFinQA \citep{chen2021finqa} score an executed final answer within tolerance; TAT-QA scores an answer span; running F1 over every figure in a free-form generation is the outlier. We extract the quarter an answer names \emph{as} its answer --- preferring the model's own emphasis, otherwise the first quarter after a superlative that is not inside a constraint clause, since ``its highest margin prior to Q2 2021 in Q3 2020'' asserts the second and merely bounds the search with the first. Rescoring the same stored answers moves the set from 0.41 to 0.58 with thirteen flips and no regressions; a fresh run under the corrected scorer reads 0.64, the remainder being generation variance.

\begin{table}[t]
\centering
\small
\caption{ECT-QA by question type under two protocols, \texttt{gemini-3.6-flash} throughout, scored on \emph{identical} generations. Strict is thresholded figure matching with the asserted-answer extraction described above. Native is the protocol of ECT-QA's authors \citep{ectqa}: an LLM judge comparing element-wise, reporting rates that sum to one per query. The ``unanswerable'' row rewards correct refusal.}
\label{tab:ectqa-types}
\begin{tabular}{@{}L{3.9cm}rrrrr@{}}
\toprule
& & \textbf{strict} & \multicolumn{3}{c}{\textbf{native, element-wise}} \\
\cmidrule(lr){4-6}
\textbf{Question type} & $n$ & \textbf{acc.} & \textbf{Correct} & \textbf{Refusal} & \textbf{Incorr.} \\
\midrule
unanswerable & 8 & \textbf{0.875} & \textbf{0.875} & 0.000 & 0.125 \\
single-time query & 21 & 0.667 & 0.761 & 0.061 & 0.179 \\
multi-time query & 31 & 0.710 & 0.830 & 0.058 & 0.112 \\
relative-time query & 8 & \textbf{1.000} & 0.938 & 0.062 & 0.000 \\
cross-company & 10 & 0.500 & 0.675 & 0.025 & 0.300 \\
\midrule
\textbf{overall} & 78 & \textbf{0.718} & \textbf{0.807} & 0.049 & 0.144 \\
\bottomrule
\end{tabular}
\end{table}

\textbf{What became of the cross-company deficit.} An earlier version of this paper reported cross-company at 0.100 and called it unexplained after falsifying three hypotheses --- retrieval coverage, refusal wording, and validity windows. It now reads 0.500 under strict scoring and 0.675 under the native protocol, and none of that movement came from the retrieval or synthesis code. The ablation of Section~\ref{sec:context} decomposes it, since the 4000-token arm is run under the current scorer on the repaired index: scoring and the missing company together account for 0.100 of the movement, and context assembly for the remaining 0.300. Three quarters of it is the assembler --- which fits, since these questions need figures from several companies at once and were therefore the most damaged by one that dropped passages. The smaller part is scoring: three of the ten have golds naming a company rather than a figure, which the discarded cosine fallback graded by vocabulary overlap.

We record this because the failure mode is instructive and cheap to repeat. Three hypotheses were tested and discarded against a number that was substantially an artefact of measurement and of a defect downstream of everything being hypothesised about. A fourth idea --- scatter-gather retrieval, querying each company's space separately and merging --- was implemented and measured before the cause was understood: it made the category \emph{worse}, and is reported here as falsified rather than omitted.

\paragraph{Comparability, and what a strict metric hides.} ECT-QA comes from TG-RAG \citep{ectqa}, whose authors do not use figure matching as their headline. They have an LLM judge perform a fine-grained element-wise comparison and report three rates summing to one per query: Correct, Refusal, Incorrect. Two differences from strict scoring matter, and both run against this system's interest under the strict metric. It is element-wise, so three of four quarters right scores 0.75 where a threshold scores zero. And a refusal is its own outcome rather than a wrong answer --- the right treatment for a corpus where some questions genuinely cannot be answered from the transcripts.

Scoring our stored answers that way gives 0.807 Correct, against 0.599 for TG-RAG and 0.405 and 0.406 for GraphRAG and LightRAG as published on their corpus slice. We hold that comparison loosely: their judge is GPT-4o-mini, which our router does not serve, and their verbatim rubric sits in an appendix truncated in the public HTML, so ours follows their description of the three categories without being their text. Judged instead by \texttt{gpt-4o}, the same answers score 0.805 --- two-tenths of a point apart across two judge families, which is the more useful fact, since the standing objection to an LLM-judged rate is that it moves with the grader.

\paragraph{A correction we found by taking the refusal rate seriously.} An earlier version of these figures put refusal at 0.173 and overall Correct at 0.677, and we read the refusal mass as a retrieval-coverage limit. It was not. Ten of the 78 questions concern a company whose sixteen transcripts are present in the corpus but had never been indexed into the realm: the indexing failed in an earlier session, and every run since had reused that realm under a \texttt{-{}-skip-index} flag that trusted it without checking. The system declined all ten questions, correctly, having nothing to answer from, and the scorer recorded ten refusals.

Indexing that company and re-running the same 78 questions against the same realm and configuration moves strict scoring from 0.641 to 0.718 and Correct from 0.677 to 0.807, while refusal falls from 0.173 to 0.049 and incorrect barely moves, 0.150 to 0.144. That is the signature of missing evidence rather than a more confident reader: ten questions became answerable and nothing else changed. The harness now counts documents per company before asking anything and refuses to run against a realm that is missing one.

We report this because the failure is a general one and the metric is what exposed it. A refusal rate is a coverage instrument, and a strict metric could not have raised the alarm, since under it a decline and a wrong answer score identically. What remains after the correction is a 0.144 incorrect rate concentrated in cross-company questions, and the two judges still disagree about how much of the residue is a decline rather than an error --- refusal reads 0.049 under \texttt{gemini-3.7-flash} and 0.127 under \texttt{gpt-4o} on identical answers --- so we report the shape rather than a point estimate.

\paragraph{Why document keying is part of the temporal design.} A corpus of sixteen quarters per company is also a stress test of identity: every transcript must resolve to its own document key, since a matching key means \emph{re-index} and therefore replacement. Section~\ref{sec:reindex} describes the composite key --- source and title together --- that guarantees this, and the dormancy rule that preserves entities whose last mentioning document is replaced. On a corpus where the same speakers discuss the same metrics sixteen times, keying by either part alone is not sufficient, and this benchmark is the instrument that establishes it: a graph that silently retained the wrong subset would answer ``unanswerable'' with complete correctness, and only a reference-scored benchmark distinguishes that from a graph that is simply cautious.

\subsection{Separating the reader from the graph}
\label{sec:readers}

Section~\ref{sec:ectqa} attributes the remaining deficit to indexing rather than to retrieval or prompting. This section measures that, using the paired arrangement of Section~\ref{sec:paired} applied to the answering model: one graph, indexed once, read by four models over the same 78 questions.

\begin{table}[t]
\centering
\small
\caption{Four answering models over one fixed graph, ECT-QA, after the corrections of Sections~\ref{sec:context} and~\ref{sec:ectqa}. Extraction was performed once by \texttt{gemini-3.6-flash}; only the model that reads the retrieved evidence and writes the answer differs. ``Refused'' counts answerable questions declined.}
\label{tab:ectqa-readers}
\begin{tabular}{@{}L{4.3cm}rrrr@{}}
\toprule
\textbf{Answering model} & \textbf{overall} & \textbf{refused} & \textbf{single} & \textbf{multi} \\
\midrule
\texttt{gemini-3.7-flash} & \textbf{0.756} & 0 & \textbf{0.71} & \textbf{0.77} \\
\texttt{gemini-3.6-flash} & 0.718 & 0 & 0.67 & 0.71 \\
\texttt{gemma-4-26b} & 0.551 & 7 & 0.62 & 0.45 \\
\texttt{gpt-oss-120b} & 0.526 & 1 & 0.43 & 0.45 \\
\midrule
\textbf{union over all four} & \textbf{0.846} & --- & --- & --- \\
\bottomrule
\end{tabular}
\end{table}

\textbf{This measurement replaces one that was an artefact, and the replacement inverts its conclusion.} The previous version of this paper reported three unrelated models scoring \emph{identically} on this benchmark --- 24 of 68 each --- and drew from that identity a graph-limited floor: 39 questions answered by no model, 57\% of the set, unreachable by any reader. Neither number survives. Re-run after the context-assembly repair of Section~\ref{sec:context} and the missing-company repair of Section~\ref{sec:ectqa}, the four readers separate across a 23-point band, and the questions no reader answers fall from 39 of 68 to \textbf{12 of 78}.

The identity was the artefact. An assembler that withheld most of the retrieved evidence starved every reader equally, and equally starved readers score alike; what looked like a property of the graph was a property of the prompt they were all being handed. We record this because the inference was reasonable on the evidence available and still wrong, and because the failure mode generalises: a measurement that makes several systems agree is as likely to indicate a shared upstream constraint as a shared limit.

\textbf{What survives is weaker and more ordinary.} Twelve questions of 78 are answered by no reader, which bounds what this graph supports at 0.846 rather than the 0.426 union reported previously. Reader choice now matters substantially --- \texttt{gemini-3.7-flash} answers 44\% more questions than \texttt{gpt-oss-120b} --- so on this corpus the answering model is a live variable rather than a rounding term, which is the opposite of what we previously concluded.

\textbf{One model remains strictly dominated.} Every question \texttt{gpt-oss-120b} answers correctly is answered correctly by at least one other reader; it contributes nothing to the union. \texttt{gemini-3.7-flash} contributes four questions no other reader solves, \texttt{gemini-3.6-flash} two, and \texttt{gemma-4-26b} one.

\textbf{Refusal behaviour separates the models more sharply than accuracy does.} \texttt{gemma-4-26b} declines seven answerable questions where both Gemini generations decline none. Willingness to commit to a partial answer is a distinct axis from the ability to find one, and a deployment that tunes only for accuracy will not see it.

\paragraph{The extractor experiment is not re-established.} The previous version also rebuilt the corpus with \texttt{gemini-3.7-flash} extraction and reported that none of the 39 unsolved questions was recovered, which supported the graph-limited reading. That experiment ran on the defective assembler, and the set it was defined against no longer exists. We have not re-run it, so we withdraw the claim rather than restate it: whether a stronger extractor recovers the remaining twelve is currently unknown. What Section~\ref{sec:models} establishes independently --- that extraction-model choice moves results more than most features under test --- is unaffected, since those measurements are relative comparisons within a single protocol.

\subsection{Traversal depth and the similarity channel}
\label{sec:retrieval-eval}

Depth and the similarity channel are evaluated on the Boeing corpus, the register where chain questions arise naturally. Table~\ref{tab:depth} reports relations retrieved per question at each depth, with the share mentioning a term from the question.

\begin{table}[t]
\centering
\small
\caption{Relations retrieved at increasing traversal depth, Boeing corpus, MiniMax-M2.7. Parenthesised figures are the share of retrieved relations mentioning a question term. Absolute on-topic signal rises with depth while precision falls sharply.}
\label{tab:depth}
\begin{tabular}{@{}L{5.2cm}rrr@{}}
\toprule
\textbf{Question} & \textbf{1 hop} & \textbf{2 hops} & \textbf{3 hops} \\
\midrule
Deferred production costs arc & 7 (100\%) & 30 (47\%) & 202 (10\%) \\
What turned cash flow negative & 13 (77\%) & 20 (60\%) & 59 (39\%) \\
737 vs.\ cash flow over time & 32 (91\%) & 125 (48\%) & 435 (23\%) \\
FAA role across filings & 31 (74\%) & 73 (52\%) & 307 (18\%) \\
\bottomrule
\end{tabular}
\end{table}

Precision collapses while absolute signal rises: three hops reaches roughly three times as many on-topic relations and ten times as much noise. Recall alone is therefore the wrong objective --- reaching more relations is trivial, reaching more \emph{relevant} ones is the claim. Depth pays exactly where the question is a chain. \emph{What caused Boeing's cash flow to turn negative} is the one question a single hop could not answer, returning ``the documents do not contain a direct statement specifically addressing'' it; at three hops the same query answers directly, connecting the revenue decline, the 737-9 grounding and the fixed-price development charges into one causal account. The other three answer at every depth.

Enabling the similarity channel raises the on-topic share of the relations that survive truncation substantially on both extraction models tested. At the default quota $q = 0.5$ it moves 49\% to 69\% on a graph built by gpt-oss-120b and 66\% to 88\% on one built by MiniMax-M2.7; at $q = 1$, where the channel has first claim on every slot, 73\% and 98\%.

\paragraph{Setting the quota requires an unbiased metric.} The scoring above counts keyword overlap with the question, which correlates with the embedding similarity the second channel ranks by. It therefore rewards the channel under test, and cannot be trusted to order intermediate quotas against $q = 1$; unsurprisingly it prefers larger $q$ monotonically. We resolve this on answers instead. For each question we generate an answer at each setting through the unmodified query path and have judge models select the better one, under three controls: the judge sees only the question and two answers labelled A and B, with no indication of provenance; no judge grades prose written by its own model; and every pair is graded twice with the labels transposed, a win counting only when the judge selects the same answer in both presentations.

The transposition control is not a formality. It discarded 7 of 20 judgements on one graph and 11 of 30 on the other --- 18 of 50 overall --- as position-determined rather than content-determined, including one judge that selected the first-presented answer almost uniformly. A single-pass protocol would have recorded every one of those as a result.

\begin{table}[t]
\centering
\small
\caption{Blind pairwise wins for $q = 0.5$ against $q = 1.0$, ten questions on two graphs built from the Boeing corpus by different extraction models, judged by \texttt{gemma-4-31B-it}, \texttt{DeepSeek-V3.2} and \texttt{gemini-3.6-flash}. Order-dependent judgements are excluded. The two graphs disagree in aggregate; they agree once questions are grouped by shape.}
\label{tab:quota}
\begin{tabular}{@{}L{6.6cm}rr@{}}
\toprule
\textbf{Question shape} & \textbf{$q = 0.5$} & \textbf{$q = 1.0$} \\
\midrule
Entity --- \emph{what role did the FAA play} & \textbf{8} & 4 \\
Thematic --- \emph{what caused cash flow to turn negative} & \textbf{7} & 4 \\
Chain --- \emph{how did regulation translate into financial consequences} & 2 & \textbf{6} \\
\bottomrule
\end{tabular}
\end{table}

Table~\ref{tab:quota} gives the outcome. Traversal-weighted retrieval leads where the question names its subject or a theme; similarity-weighted retrieval leads on chain questions, and led 5--0 on the weaker of the two graphs, consistent with traversal accumulating noise faster with depth when extraction is poorer. The judges' written justifications explain why no setting dominates: both directions are argued on identical grounds --- concrete figures present, extraneous material absent --- with each setting praised for citing specific charges in different comparisons. The criterion is stable across judges; which setting satisfies it is a property of the graph. We therefore report $q$ as a parameter with a default of $0.5$ rather than a solved value, and ship the protocol above so it can be measured per corpus.

\subsection{Paired ablation, and the re-indexing noise floor}
\label{sec:paired}

The three mechanisms of Section~\ref{sec:diversification} were evaluated on a 20-instance LongMemEval slice. How that evaluation is arranged turns out to matter more than the mechanisms do, and the arrangement is the result we would most like the area to adopt.

\begin{table}[t]
\centering
\small
\caption{Paired ablation on a 20-instance LongMemEval slice. Each instance is indexed \emph{once}, and every variant answers from that identical graph. Contradiction detection changes what is written rather than what is read, so it cannot share a graph with the others and carries its own re-indexed baseline. These are relative comparisons on a small sample; the absolute values are not comparable to Table~\ref{tab:longmemeval}.}
\label{tab:paired}
\begin{tabular}{@{}L{7.2cm}rr@{}}
\toprule
& \textbf{accuracy} & \textbf{delta} \\
\midrule
baseline & 60\% & --- \\
\quad + MMR & 65\% & $+5$ \\
\quad + node-distance reranking & \textbf{70\%} & $\mathbf{+10}$ \\
\midrule
contradiction detection (own baseline, 70\%) & 75\% & $+5$ \\
\bottomrule
\end{tabular}
\end{table}

\textbf{Measured the obvious way, node-distance reranking scores $-5$. The same code scores $+10$ once every variant reads the same graph.} The difference is entirely in whether each arm re-indexes. Repeated runs explain why: with the graph held fixed the variants agreed in 80 of 80 cells, so the read path is effectively deterministic and all the movement lives in extraction. An ablation that rebuilds the index per arm is therefore not measuring the feature. It is measuring the extraction model, with the feature as a rounding term.

The magnitude is not marginal. \textbf{The same baseline configuration has scored 75\%, 70\% and 60\% on identical code and identical instances, purely from re-indexing.} That 15-point band is wider than any retrieval effect in Table~\ref{tab:paired}, and it is the reason those effects are only detectable when it is eliminated by construction.

The point generalises past our own features, and we found it by being caught out. Two indexing-side changes were built to address the diagnosed knowledge-update and temporal-reasoning deficits of Section~\ref{sec:longmemeval}: enabling LLM contradiction detection, and broadening extraction to mundane dated actions. The changes are unrelated. They produced near-identical category swings --- knowledge-update down 16 to 21 points, preference up 18, in both. That is not two features failing the same way. It is re-indexing reshuffling the categories by more than either effect being hunted, on a benchmark that rebuilds every instance's graph per run.

Two consequences follow, and we state them as costs rather than as findings. Read-side changes can be evaluated cheaply, because pairing on a fixed graph removes the dominant variance term. \textbf{Index-side changes cannot be evaluated at 120 instances at all}: the honest price of testing one is repeated indexing per configuration, or the full benchmark per candidate. Both proposed fixes therefore remain untested rather than rejected, and the two diagnoses remain open.

We are correspondingly cautious about Table~\ref{tab:paired} itself. At 20 instances a sign test on node-distance reranking gives $p = 0.31$; separating effects of this size cleanly needs 100 to 200. All three point the same way and each addresses a mechanism-level gap we can name --- MMR because reciprocal rank fusion says nothing about redundancy between channels that correlate, node distance because two of the three channels never walked and so report every result at one hop, contradiction detection because declared exclusive groups cannot see an object change under a fixed predicate. That is why all three ship, and why all three ship off by default.

\subsection{Clustering and report quality}

\begin{figure}[t]
  \centering
  \includegraphics[width=0.98\linewidth]{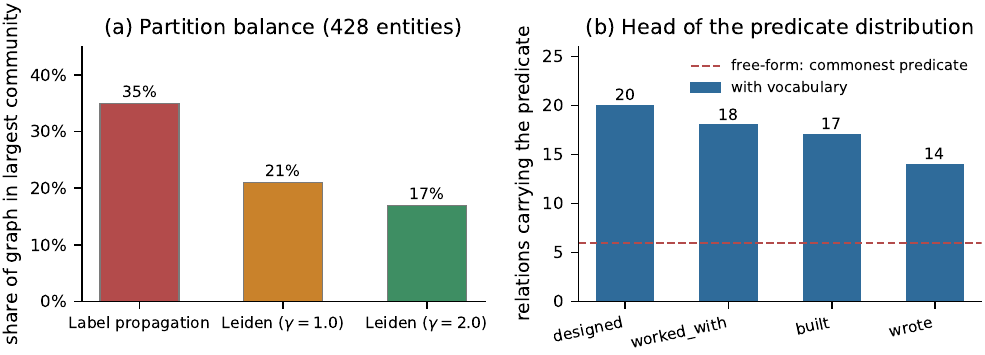}
  \caption{(a) Share of the 428-entity graph held by the largest detected community. Label propagation, the dependency-free fallback, produces a cluster covering a third of the corpus; Leiden at $\gamma = 2.0$ halves it. (b) Head of the predicate frequency distribution under the \texttt{biography} vocabulary, against the free-form baseline in which the commonest predicate over the whole graph appears six times (dashed).}
  \label{fig:partition}
\end{figure}

Partition balance depends heavily on the detector (Figure~\ref{fig:partition}(a)). On the 428-entity Wikipedia graph, the largest community holds 35\% of the graph under label propagation, 21\% under Leiden at $\gamma = 1.0$, and 17\% at $\gamma = 2.0$. A community covering a third of the graph is not a theme --- it is a summary of everything, and it will be retrieved for every global question because it is topically adjacent to all of them. This is why Leiden ships as a core dependency and the deterministic fallback exists only for environments where the native build is unavailable. At $\gamma = 2.0$ the corpus yields 28 detected communities, of which the 12 largest are summarised.

Report quality is likewise model-dependent in a way that matters for how reports can be used. On the same graph with the same clustering, Llama-3.3-70B produced reports in 20\,s in which every community received an importance rating of exactly 8.0, with two clusters given identical titles; stronger models produced three to six distinct rating values with no duplicate titles and findings naming specific mechanisms rather than restating the cluster's subject. The \texttt{rating} field is only meaningful with a model that discriminates, so a deployment that ranks or filters communities by rating should verify that it varies. Title collisions are handled structurally by the disambiguation rule of Section~\ref{sec:communities}, which is why they are cosmetic rather than a retrieval problem.

\section{Limitations}
\label{sec:limitations}

\textbf{Five registers, one language.} Wikipedia, nineteenth-century French fiction in translation, US financial filings, earnings call transcripts and multi-session chat histories cover more ground than a single-domain evaluation, and the last two are the conversational and spoken registers a previous version of this paper named as untested. All are English. We would still expect meaningfully different behaviour on code, on tabular or semi-structured sources, or on corpora whose significant entities are not named-entity-like. Predicate vocabularies are domain-specific by construction: the mechanism generalises, the vocabularies do not.

\textbf{The evaluations are complementary, and none is complete on its own.} LongMemEval scores answers against references but compares against published figures we did not produce; ECT-QA scores answers against published figures for three systems, but under a rubric we reconstructed rather than theirs and on a different corpus slice, so the margin there is approximate in a way the LongMemEval comparison is not; the corpus comparisons hold the extraction model, embedding model, chunker and synthesis prompt fixed across both systems but score no answers. The combination we would most want --- reference-scored answers with the extraction model pinned across systems --- requires running the baseline ourselves under our models, which for Zep means standing up a service we do not operate. Section~\ref{sec:models} shows that varying the extraction model alone moves results enough to dominate any system-level difference, so the benchmark comparison should be read as informative about the model tier rather than as a controlled system-level result.

\textbf{Benchmark comparability.} The Zep figures are as published, and the model comparison is uncontrolled in a direction we cannot sign: their \texttt{gpt-4o} was the frontier tier of its generation where our flash-class models are the low-cost tier of theirs, but roughly two years separate the generations, and a newer small model beating an older frontier one is routine. We therefore claim neither that the model gap favours them nor that it favours us. Their single-judge \texttt{gpt-4o} protocol likewise differs from our majority panels in kind rather than in strength. One instance of 500 is excluded, having been refused by both extraction prompts; the harness marks a run non-reportable until that count is zero, so it is stated rather than absorbed. We report a single full-benchmark run: the noise floor below applies to it as much as to the ablations, and we have not characterised its width at $n = 499$.

\textbf{What the remaining ECT-QA deficit is has not been established.} Twelve of 78 questions are answered by none of four readers over a fixed graph, which bounds what this graph supports rather than explaining what is missing from it. The previous version of this paper attributed a much larger remainder to extraction and supported that with a rebuild under a different extraction model; both the remainder and the rebuild belonged to a defective engine, so the attribution is withdrawn rather than restated. We do not currently know whether those twelve are limited by extraction, by retrieval, or by evidence the transcripts do not contain, and we have not re-run the extractor substitution that would begin to separate those.

\textbf{ECT-QA comparability is approximate, and what remains is unexplained.} Scoring under the authors' own protocol puts this system beside their published figures, but not exactly: their judge model is unavailable to us and their verbatim rubric is truncated in the public HTML, so ours reproduces their described categories rather than their text. A margin of a few points against those baselines should not be read as decisive. Within our own numbers, what remains after the correction described in Section~\ref{sec:ectqa} is a 0.144 incorrect rate concentrated in cross-company questions, against a refusal rate of 0.049 to 0.127 depending on the judge. We have not established what the cross-company residue is. The ECT-QA figures also share one index across the reader comparison, which is what makes those arms comparable to each other but leaves the absolute level resting on one indexing pass --- and that single pass is precisely what failed silently for one company until the refusal rate exposed it.

\textbf{An LLM judge panel was our least reliable instrument.} Section~\ref{sec:ectqa} reports a panel failing an answer with three exact figure matches and a fourth within 0.3 points, and scoring six points below a deterministic metric over identical answers. We draw a narrow conclusion, since our panel was three mid-tier models on numeric answers: where reference answers admit a deterministic comparison, a judge is a worse instrument than the comparison, and the gap is large enough to be mistaken for a system result. Our LongMemEval figures use a judge panel because that benchmark's answers are prose and admit no such comparison, which is a caveat on those figures rather than an exception to this one.

\textbf{Index-side changes are below our measurement floor.} Section~\ref{sec:paired} quantifies this: re-indexing alone moves the same configuration across a 15-point band, wider than any effect we tested. Pairing on a fixed graph removes it for read-side features, but no such device exists for a change that alters what is written. Two proposed fixes for the diagnosed knowledge-update and temporal-reasoning deficits are therefore untested rather than rejected, and we do not know how many previously reported Graph RAG ablations --- including our own earlier ones --- would survive being re-run paired.

\textbf{Most mechanism ablations predate the context-assembly fix.} Section~\ref{sec:context} raised every end-to-end figure by five to nine points, and the per-mechanism ablations, the extraction-model sweep and the reader comparison were all measured before it under the defective assembler. We have re-run the end-to-end benchmarks and report those; we have not re-run every ablation. Their \emph{directions} are what those sections claim and we expect those to hold, since the assembler affected all arms of a paired comparison alike, but their absolute magnitudes belong to the old engine and a reader should treat them as such. Where a number in this paper is post-fix we say so; where it is not, it is not.

\textbf{Belief time does not govern entity descriptions.} \texttt{as\_believed\_at} filters relations and, since this revision, the document channel; entity descriptions are living summaries rewritten whenever a later document mentions the entity, and carry no per-instant version. A belief-time query can therefore surface a description edited after the instant it asks about, even though every figure and passage it sees is correctly filtered. Versioning descriptions is the fix and is not implemented. We state the boundary because the guarantee is otherwise easy to over-read.

\textbf{Single runs.} Each configuration was indexed once. LLM extraction is stochastic, so smaller differences --- 428 against 447 entities, 1{,}377 against 1{,}246 --- are within plausible run-to-run variance and should not be read as precise. The large, categorical effects are the ones we would defend: zero against six fragmented vertices; zero against 21 negations; 12\% against 44\% vocabulary coverage; 133\% against 58\% label ratio; 13 and 8 supersessions against zero.

\textbf{No cross-system throughput claim.} We report no indexing-throughput comparison against LightRAG, for the reason given in Section~\ref{sec:lightrag}: a transactional database behind a connection and an in-process file-backed graph do not offer the same guarantees, so the ratio is not attributable to the pipeline. A separate confound would apply to the Boeing corpus regardless --- both runs there shared a router with other work.

\textbf{Supersession requires density and a declared vocabulary.} It fires only when the same entity pair is characterised twice by predicates in a declared exclusive group. A sparse index fires none, and a deployment that declares no groups gets none. Both are visible in the reported counts (13 and 8) being small in absolute terms: they are the relationships that genuinely reversed, not a large fraction of the graph.

\textbf{Fail-closed against malformed output, not omitted output.} The gates reject structure that is present and wrong. Nothing detects structure that is absent: a model that never emits aliases, or never sets the negation flag, produces a graph that is internally consistent, passes every gate, and is quietly missing a capability. Section~\ref{sec:models} is the evidence, and we know of no way to detect it from the stored graph alone without a reference extraction.

\textbf{Traversal depth is bounded by a fixed budget, not by relevance.} Depth is a per-query constant with a per-entity edge cap, and hop-major ordering is what keeps deeper walks from degrading (Section~\ref{sec:retrieval-eval}). Neither is a relevance-pruning policy: nothing decides \emph{which} branch is worth another hop, so precision still falls with depth (100\% to 10\% on one question) and the budget rather than the graph decides where the walk stops. A path-based method such as personalised PageRank \citep{gutierrez2024hipporag} would prune by relevance instead, and we have not compared against one.

\textbf{The retrieval quota is a parameter, not a solved value.} This limitation applies to the fixed-share interleave, which reciprocal rank fusion replaced as the default in version 1.6.0 precisely because it needs no such constant; the quota is retained as an option and the caveat travels with it. Enabling similarity-based relation retrieval is unambiguous on both models tested. How much of the truncated context it should claim is not: two graphs built from the same corpus by different extraction models pointed in opposite directions in aggregate, agreeing only once questions were grouped by shape (Table~\ref{tab:quota}). Ten questions and two graphs is a thin basis for a default, and 18 of 50 judgements were discarded as order-dependent, which further reduces the effective sample. We report the protocol rather than a recommended value.

\textbf{Index-time cost.} Gleaning doubles LLM calls per chunk, community construction costs one call per summarised cluster, and ordered graph writes cost throughput. These are amortised over all subsequent queries, but they scale with corpus size rather than with query volume.

\section{Conclusion}

\texttt{post-graph-rag} shows that a Graph RAG system does not require a federation of specialised stores: text chunks with HNSW-indexed embeddings, a canonical entity graph, document-to-entity mention edges and community summaries all sit comfortably in one PostgreSQL database, with tenancy expressed as a schema-per-realm partition and space scoping applied uniformly to search and traversal.

Two design commitments account for most of the measured difference. The first is treating extraction output as untrusted input: refusing to write placeholder structure, refusing vague predicates, refusing pronominal and conjunctive names and bare quantities, refusing to invert a predicate to express a denial. Each mechanism is doing identifiable work --- gleaning buys 40\% more relations, alias resolution eliminates entity fragmentation, corroboration weighting eliminates duplicate edges, and a controlled vocabulary converts an unqueryable long tail of one-off predicates into a distribution with a usable head. Against LightRAG under an identical extraction model, this yields a denser graph on all three corpora, a far more queryable one, lower query latency and comparable answer quality, at a one-off index-time cost, since ordering graph writes is what keeps entity resolution correct under concurrency and is what the density is bought with.

The second is modelling time. A graph that only accumulates asserts that everything in it holds at once, which is false of any corpus whose facts move. Recording validity where the prose states it, and closing incompatible relations from document order where it does not, is enough to make a graph built from three novels or a decade of filings answer questions about \emph{trajectories} rather than about a flattened present. That capability has no counterpart in the baseline we compared against.

End-to-end, the architecture leads on a public benchmark it was not designed around. On LongMemEval it scores 94.0\% driving a flash-class model, ahead of the strongest published configuration on all six question types and by 22.8 points overall, with its widest margins on temporal reasoning ($+33.8$) and multi-session synthesis ($+32.3$) --- the two categories a bi-temporal graph exists to serve.

The largest single contribution is temporal grounding in the prompt. Extracting validity, normalising it and indexing it is necessary but not sufficient; carrying it through to synthesis is what makes ordering and duration questions answerable, and ablating that step costs 38.4 points on temporal reasoning. Two index-side candidates aimed at the same gap moved nothing, which places the mechanism on the read side and suggests the render path deserves treating as part of a temporal design rather than as formatting. On ECT-QA, a harder register in which every metric carries sixteen competing values, it reaches 0.718 under strict figure matching and 0.807 under that benchmark's own element-wise protocol, with correct refusal on 0.875 of the deliberately unanswerable questions. Cross-company remains the weakest category at 0.500, and what limits the twelve questions no reader answers is not established.

On ECT-QA the same paired arrangement was applied to the answering model, and it corrected us. An earlier version of this work found four readers scoring identically over one fixed graph and inferred a graph-limited floor beneath them --- 57\% of the set beyond any reader's reach. Repaired and re-run, the readers separate across 23 points and the unreachable remainder falls to 15\%. The identity had been an artefact of an assembler that starved every reader of the same evidence, and equally starved readers score alike. The paired sweep was the right instrument and it gave the wrong answer, because the quantity it measures --- what a reader can recover from the prompt it is handed --- is only a statement about the graph when the prompt faithfully carries what the graph holds.

The measurements also carry three warnings for the field, and the first two are about measurement rather than architecture. The first is that the instrument is often the thing being measured. Most of the changes that moved our ECT-QA score were repairs to the evaluation or to the harness rather than to the system --- a judge panel that failed arithmetically correct answers, a similarity metric that could not separate correct answers from refusals, an answer prompt that instructed the refusals it was then penalised for, a period metric that scored a superlative answer against its own justification, and a flag that reused an index without checking it contained the companies about to be questioned. A system evaluated once, with none of those repairs made, would have been reported at 0.162 rather than 0.718, and most of that difference is in the measurement and the harness rather than in the system.

The second is closely related: on a benchmark that rebuilds its graph per run, re-indexing moved the same configuration across a 15-point band, and a retrieval feature that scores $-5$ when each arm re-indexes scores $+10$ when every arm reads one fixed graph. Ablations arranged the first way are measuring the extraction model. Pairing on a held-fixed graph is cheap, and we would encourage it as the default arrangement; the harder consequence is that index-side changes have no equivalent device and are correspondingly expensive to evaluate honestly.

The third is the older one, and it is about what gets measured rather than how. Holding everything else fixed and changing only the extraction model tripled alias emission, moved negation capture from zero to 21 relations, and moved vocabulary adherence from 33\% to 94\%. Graph RAG results reported without pinning the extraction model are not reproducible, and comparisons between systems evaluated under different extraction models are not comparisons between the systems. A shared protocol that pins the extraction model, the embedding model and the chunker --- and that measures graph structure alongside answer quality --- would do more for the area than another retrieval topology.

\subsubsection*{Reproducibility}

The implementation is Apache 2.0 licensed at \url{https://github.com/crajah/post-graph-rag} and published on PyPI as \texttt{post-graph-rag}; the graph substrate it builds on is at \url{https://github.com/crajah/post-graph} (PyPI: \texttt{post-graph}). The evaluation harness --- corpus fetch, SEC EDGAR download, indexing, graph analysis, cross-run comparison, the LightRAG bridge, the temporal evaluation and the LongMemEval runner, including the paired-ablation mode that holds one index fixed across variants --- is in the \texttt{evaluation/} directory; \texttt{evaluation\_notes.md} in this paper's repository records the exact commands, model line-up and raw tables behind every number in Section~\ref{sec:evaluation}. Figures are regenerated from those transcribed measurements by \texttt{scripts/generate\_paper\_figures.py} with no network or database access.

\end{document}